\documentclass[preprint,3p,times,twocolumn,authoryear]{elsarticle}

\usepackage{amssymb}
\usepackage{amsmath}
\usepackage{booktabs}
\usepackage{multirow}
\usepackage{xcolor}
\usepackage{amssymb}
\usepackage{pifont}
\usepackage{bm}

\definecolor{deepred}{HTML}{B22222}

\usepackage[skins, breakable, listings, theorems]{tcolorbox}
\usepackage{array}
\usepackage{tabularx}
\usepackage{ragged2e}

\usepackage{booktabs}
\usepackage{array}
\usepackage{multirow}
\usepackage{makecell}
\usepackage{threeparttable}

\def\sC{{\mathbb{C}}}
\def\sD{{\mathbb{D}}}

\def\sT{{\mathbb{T}}}

\journal{arXiv}

\begin{document}

\begin{frontmatter}


%%     addressline={},
%%     city={},
%%     postcode={},
%%     state={},
%%     country={}}

\title{PathFound: An Agentic Multimodal Model Activating Evidence-seeking Pathological Diagnosis}

%%     addressline={},
%%     city={},
%%     postcode={},
%%     state={},
%%     country={}}
%%     addressline={},
%%     city={},
%%     postcode={},
%%     state={},
%%     country={}}

\author[1]{Shengyi~Hua\fnref{cofirst}}
\author[2]{Jianfeng~Wu\fnref{cofirst}}
\author[1]{Tianle~Shen}
\author[1]{Kangzhe~Hu}
\author[1]{Zhongzhen~Huang}
\author[5]{Shujuan~Ni}
\author[6]{Zhihong~Zhang}
\author[5,8,9]{Yuan~Li}
\author[2,3,4]{Zhe~Wang\corref{cor1}}
\author[1,7]{Xiaofan~Zhang\corref{cor1}}

\fntext[cofirst]{These authors contributed equally to this work.}
\cortext[cor1]{Corresponding authors: Zhe Wang (zhwang@fmmu.edu.cn), and Xiaofan Zhang (xiaofan.zhang@sjtu.edu.cn).}

\affiliation[1]{organization={Qing Yuan Research Institute, Shanghai Jiao Tong University},%Department and Organization
    city={Shanghai},
    postcode={200240}, 
    state={Shanghai},
    country={China}}
\affiliation[2]{organization={State Key Laboratory of Holistic Integrative Management of Gastrointestinal Cancers, Department of Pathology, School of Basic Medicine and Xijing Hospital, Fourth Military Medical University},%Department and Organization
    city={Xi'an},
    postcode={710032}, 
    state={Shaanxi},
    country={China}}
\affiliation[7]{organization={Shanghai Innovation Institute},%Department and Organization
    city={Shanghai},
    postcode={200231}, 
    state={Shanghai},
    country={China}}
\affiliation[3]{organization={Department of Pathology, The First Affiliated Hospital
of USTC, Division of Life Sciences and Medicine, University of Science and Technology of China},%Department and Organization
    city={Hefei},
    postcode={230036}, 
    state={Anhui},
    country={China}}
\affiliation[4]{organization={Intelligent Pathology Institute, Division of Life
Sciences and Medicine},%Department and Organization
    city={Hefei},
    postcode={230036}, 
    state={Anhui},
    country={China}}
\affiliation[5]{organization={Department of Pathology, Fudan University Shanghai Cancer Center},%Department and Organization
    city={Shanghai},
    postcode={200032}, 
    state={Shanghai},
    country={China}}
\affiliation[8]{organization={Department of Oncology, Shanghai Medical College, Fudan University},%Department and Organization
    city={Shanghai},
    postcode={200032}, 
    state={Shanghai},
    country={China}}
\affiliation[9]{organization={Institute of Pathology, Fudan University},%Department and Organization
    city={Shanghai},
    postcode={200032}, 
    state={Shanghai},
    country={China}}
\affiliation[6]{organization={Department of Pathology, The First Affiliated Hospital with Nanjing Medical University},%Department and Organization
    city={Nanjing},
    postcode={210029}, 
    state={Jiangsu},
    country={China}}

%% Abstract
\begin{abstract}
Recent pathological foundation models have substantially advanced visual representation learning and multimodal interaction. However, most models still rely on a static inference paradigm in which whole-slide images are processed once to produce predictions, without reassessment or targeted evidence acquisition under ambiguous diagnoses. This contrasts with clinical diagnostic workflows that refine hypotheses through repeated slide observations and further examination requests. We propose PathFound, an agentic multimodal model designed to support evidence-seeking inference in pathological diagnosis. PathFound integrates the power of pathological visual foundation models, vision-language models, and reasoning models trained with reinforcement learning to perform proactive information acquisition and diagnosis refinement by progressing through the initial diagnosis, evidence-seeking, and final decision stages. Across several large multimodal models, adopting this strategy consistently improves diagnostic accuracy, indicating the effectiveness of evidence-seeking workflows in computational pathology. Among these models, PathFound achieves state-of-the-art diagnostic performance across diverse clinical scenarios and demonstrates strong potential to discover subtle details, such as nuclear features and local invasions.
\end{abstract}

% \includegraphics[width=\textwidth]{graphic_abs.pdf}

% \begin{highlights}
% \item We propose PathFound, a large agentic multimodal model for pathological diagnosis.  
% \item Evidence-seeking diagnosis is empirically superior to one-pass diagnosis.
% \item PathFound achieves state-of-the-art performance across different scenarios.
% \item The slide highlighter and vision interpreter can provide decent visual understanding.
% \item Trained with RLVR, the diagnostic reasoner can offer complex and plausible diagnoses.
% \end{highlights}

%% Keywords

\end{frontmatter}




\section{Introduction}
Recent advances in pathological foundation models have substantially reshaped computational pathology. This progress can be broadly characterized by two stages: early self-supervised visual foundation models (VFMs) that learn rich morphological representations from whole-slide images (WSIs)~\citep{gigapath, chief} and cropped patches~\citep{uni, virchow, virchow2}, and more recent vision-language models (VLMs) that enable increasingly flexible communication with users. 

Models~\citep{plip, conch} pretrained with contrastive language-image objectives, such as CLIP~\citep{clip} and CoCa~\citep{coca}, exhibit promising zero-shot generalization to unseen categories. More recent pathological multimodal models serve as copilots~\citep{pathchat, cpathomni}, further extending the boundary by enabling conversational interactions and supporting diverse diagnostic-related tasks. To improve WSI analysis and pathological diagnosis, some copilots~\citep{pathfinder, slideseek, cpathagent} incorporate navigation or planning agents that iteratively select informative regions for inspection, forming an inner loop of slide observation that refines visual perception through repeated region proposal and analysis. 

\begin{figure*}[ht]
  \centering
  \includegraphics[width=\linewidth]{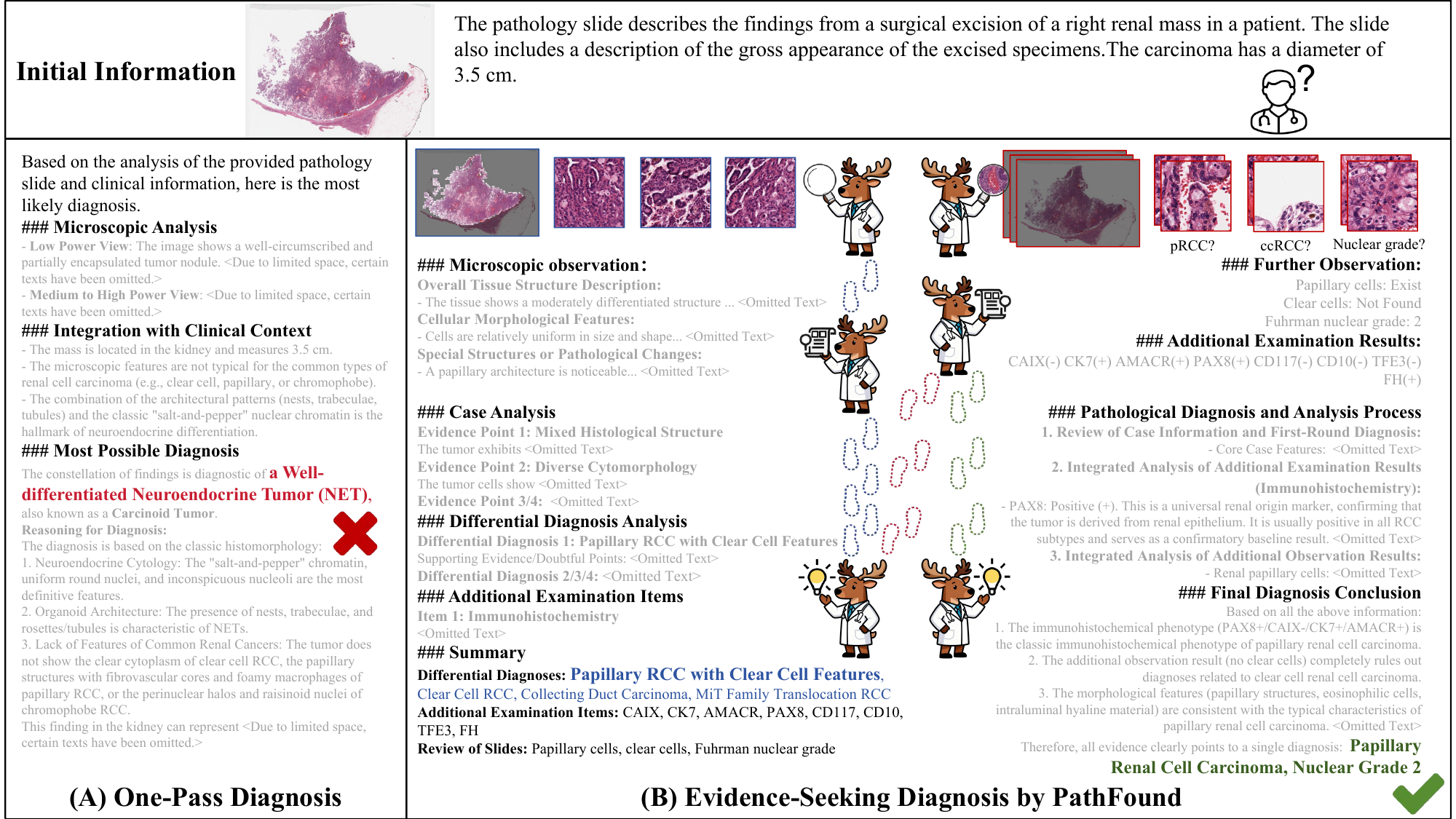}
  \caption {Comparison between conventional one-pass diagnosis and PathFound’s evidence-seeking diagnostic paradigm. (A) One-pass diagnosis: existing multimodal models analyze a whole-slide image once and generate a final prediction without revisiting the slide or refining the diagnostic hypothesis. (B) Evidence-seeking diagnosis with PathFound: the diagnosis proceeds through an outer loop of hypothesis formulation, targeted evidence acquisition (via slide re-observation or external tests), and conclusion refinement, enabling progressive, uncertainty-aware diagnostic reasoning.}
  \label{fig:vis_case}
\end{figure*}

Despite these advances, a fundamental gap remains between current multimodal models and real-world clinical diagnostic workflows. Most existing systems still operate under a ``read-once, predict-once'' paradigm, in which a WSI is analyzed once to directly produce a final answer, as illustrated in Fig.~\ref{fig:vis_case}(A). Even when iterative navigation is introduced, slide assessment primarily serves to optimize visual understanding for a fixed prediction objective, and the diagnostic conclusion itself is neither revisited nor revised. 

In contrast, routine pathological diagnosis is inherently progressive and hypothesis-driven. Pathologists typically begin with a global assessment of the slide to establish an initial diagnostic hypothesis (e.g., suspecting renal cell carcinoma). This hypothesis then guides subsequent actions, including targeted re-observation of specific regions to assess fine-grained features (such as nuclear grade) and consultation of external evidence, such as immunohistochemistry (IHC) results, when ambiguity remains. Diagnosis is refined through repeated cycles of evidence gathering and hypothesis updating. Current models lack the ability to proactively revisit WSIs under different purposes, seek targeted evidence, or gradually refine conclusions when diagnostic uncertainty persists. 

To bridge this gap, we propose PathFound, a large agentic multimodal model designed to align pathological diagnosis with clinical reasoning. As shown in Fig.~\ref{fig:vis_case}(B), rather than passively answering questions from static inputs, PathFound iteratively formulates diagnostic hypotheses, actively acquires visual or external evidence, and refines its conclusions until a precise diagnosis is achieved. Contrary to previous agentic models that focus on slide navigation, PathFound elevates slide re-observation from an inner perceptual optimization step to an integral component of diagnostic reasoning, forming an outer loop that spans hypothesis formulation, evidence acquisition, and conclusion refinement. 

PathFound integrates three complementary components. A slide highlighter, which is built upon pathological VFMs, condenses large WSIs into representative regions of interest (RoIs). A vision interpreter, adapted from general VLMs, then translates these RoIs into textual observations. A diagnostic reasoner, trained with reinforcement learning with verifiable rewards (RLVR)~\citep{deepseekai2025deepseekr1incentivizingreasoningcapability}, orchestrates evidence acquisition, interacts with users, and manages the overall diagnostic process. During a diagnostic session, PathFound follows a structured three-stage protocol to trigger the three modules, mirroring pathologists' coarse-to-fine reasoning. The protocol begins with an exploratory stage involving all three modules to form initial diagnostic hypotheses and identify informative queries for additional evidence. With limited information in hand, it proceeds to an evidence-seeking stage that actively acquires targeted visual information by re-triggering the slide highlighter and vision interpreter, and by obtaining clinical results from external requests. Once sufficient evidence has been collected, the diagnostic reasoner concludes with a decision stage that consolidates the gathered evidence to produce a final diagnosis. This iterative design enables progressive hypothesis refinement rather than relying on a single, static inference. The code will be made public at https://github.com/hsymm/PathFound.

Our contributions are threefold. (1) We introduce PathFound, an agentic pathological multimodal model that performs progressive, evidence-seeking diagnosis aligned with clinical practice. (2) We present a unified framework integrating slide highlighting, visual interpretation, and diagnostic reasoning, leveraging pathological foundation models and RLVR-trained reasoning models. (3) We empirically demonstrate that evidence-seeking diagnostic reasoning significantly improves performance across multiple large VLM architectures and clinical tasks. PathFound achieves state-of-the-art results on most evaluation metrics, including diagnostic accuracy and performance in discovering subtle details, such as nuclear characteristics and invasion patterns.

\section{Related Work}

\subsection{Vision Foundation Models in Pathology}
Early work on vision encoders laid the foundation for modern pathology models~\citep{ctranspath, pathoduet}. UNI, regarded as the first large-scale pathological model introduced by~\citet{uni}, is a self-supervised ViT trained on over 100 million image patches from 100,000 H\&E WSIs across 20 tissue types, and was evaluated on 34 downstream tasks. Likewise, Virchow~\citep{virchow} is a 632M-parameter ViT pretrained with DINOv2~\citep{dinov2} on about 1.5M WSIs, demonstrating outstanding performance in pan-cancer detection and rare cancer classification. These models showed that scaling unlabeled histology data yields strong, generalizable features for diverse tasks. These two models later proposed their larger versions, UNI2~\citep{uni} and Virchow2~\citep{virchow2}, which required more resources. More recent models have emphasized downstream task evaluation. For instance, GPFM~\citep{gpfm} uses unified knowledge distillation across over 72,000 WSIs from 34 tissues and achieves superior performance on 72 pathology tasks. PathOrchestra~\citep{pathorchestra} is trained via self-supervision on 300K whole slides from 20 tissue types and evaluated on 112 tasks. In addition to models that focus on patch features, a series of works learns slide-level features. \citet{gigapath} introduced the GigaPath with LongNet architecture for gigapixel slides, pretrained on 1.3 billion tiles from 171,189 WSIs. CHIEF~\citep{chief} exploited a weakly-supervised method to extract pathology imaging features from slides. \citet{ma2025pathbenchcomprehensivecomparisonbenchmark} benchmarked the performance of existing pathological VFMs and found that model performance is task-related and different models specialize in different task scopes. 

\subsection{Multimodal Foundation Models in Pathology}
Multimodal pathology models have similarly evolved. Early vision-language models like PLIP~\citep{plip} and CONCH~\citep{conch} leveraged CLIP-style contrastive pretraining like CLIP and CoCa. PLIP was trained on 200K pathology images paired with Twitter text, enabling CLIP-like image-text alignment for zero-shot classification and retrieval. CONCH was trained on 1.17 million curated image-caption pairs and evaluated on 14 pathology benchmarks. MUSK~\citep{musk} is another CLIP-based model, but warmed up with unified masked modeling pretrained on 50M pathology images and 1B pathology-related text tokens. PRISM~\citep{prism} is a vision-language encoder-decoder for slide-level tasks, built on slide embeddings and 195K clinical reports. TITAN~\citep{titan} is also a slide-level ViT-based encoder pretrained on 335K WSIs with 182K real and 423K synthetic captions, achieving SOTA slide embeddings for classification, retrieval, and report generation. mSTAR~\citep{xu2025multimodalknowledgeenhancedwholeslidepathology} further incorporates molecular-level gene expression data as an additional modality, in addition to visual slides and textual reports. These models use CLIP/CoCa-like architectures to embed images and text jointly, but generally produce only fixed-form outputs rather than flexible free text.

\subsection{Multimodal Chat Models in Pathology}
More recently, researchers have moved toward generative LLM-based systems. PathChat~\citep{pathchat} is a pathology copilot that connects a vision encoder to a pretrained language model and then fine-tunes on 456K visual-question-answer pairs. It supports multi-turn conversations and achieves SOTA accuracy on its diagnostic tasks. CPath-Omni~\citep{cpathomni} is a 15B-parameter VLM, integrating a pathology-specific CLIP (CPath-CLIP) into an LLM (Qwen2.5-14B), instruction-tuned on diverse patch- and WSI-level tasks. Patho-R1~\citep{zhang2025pathor1multimodalreinforcementlearningbased} and SmartPath-R1~\citep{xu2025versatilepathologycopilotreasoning} both leverage a combined training scope of supervised finetuning and reinforcement learning to support thoughtful reasoning over pathological tasks. Agentic multi-stage frameworks have also emerged. For example, PathFinder~\citep{pathfinder} leverages a specialized dataset of pathologists' viewports and uses four collaborative agents (Triage, Navigation, Description, Diagnosis) to simulate the viewport shifts during slide observation. SlideSeek~\citep{slideseek} autonomously manages slide analysis in collaboration with a powerful supervisor agent (OpenAI o1), numerous explorer agents (GPT-4o), and an updated PathChat+ dealing with diagnoses. CPathAgent~\citep{cpathagent} similarly mimics diagnostic reasoning by alternating low- and high-magnification views, zooming and panning to gather evidence, and generating explanations, aiming for a more interpretable, stepwise rationale. Pathology-CoT~\citep{wang2025pathologycotlearningvisualchainofthought} further extends the navigation loop with behavior-grounded interpretability. These LLM-driven and agentic models show promise, but most still focus on providing thoughtful, precise reasoning in a single assessment of slides with a fixed purpose rather than on a dynamic refinement process involving an outer diagnostic reasoning loop of evidence acquisition.

\begin{figure*}[ht]
  \centering
  \includegraphics[width=\linewidth]{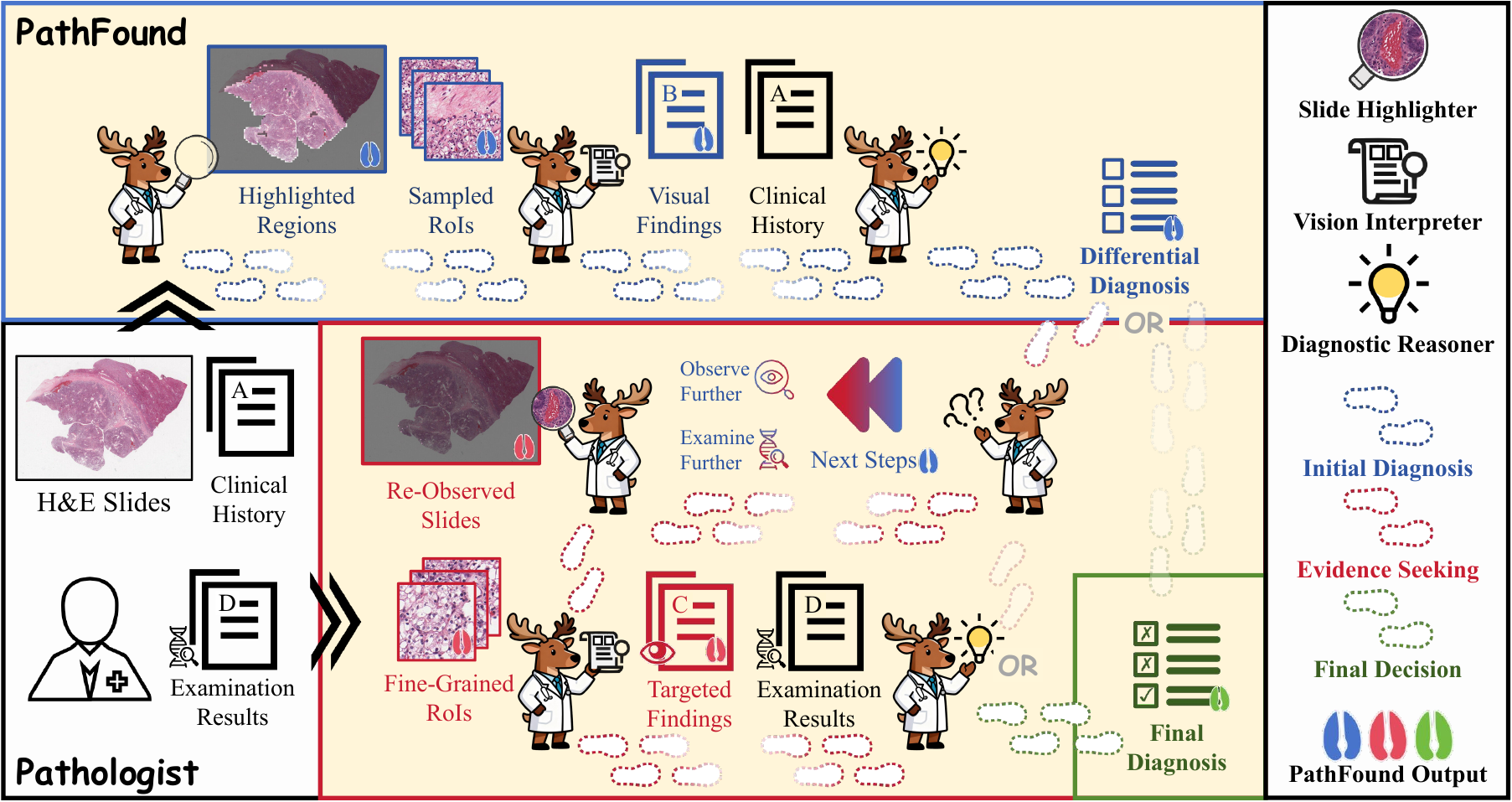}
  \caption {A complete diagnostic \textit{path} found by PathFound. The path starts with an initial diagnosis stage, where it generates a list of possible diagnoses and plans the next steps based on preliminary information. Then it transitions into an evidence-seeking mode, re-observing the slides with specific purposes and collecting external examination results. It ends with a final decision stage, in which a precise diagnosis is derived with the additional evidence. Notably, there may be more routes for practical use.}
  \label{fig:overview}
\end{figure*}

\section{Method}
We introduce \textbf{PathFound}, an agentic multimodal framework designed for evidence-seeking diagnosis in pathology. Rather than performing a single-pass prediction, PathFound frames diagnosis as an iterative, multi-turn process that integrates perception, reasoning, and targeted information retrieval. As illustrated in Fig.~\ref{fig:overview}, the framework comprises three complementary modules, the slide highlighter, the vision interpreter, and the diagnostic reasoner, and uses a three-stage protocol. Below, we first detail the multi-turn agentic process of pathological diagnosis (Section~\ref{sec:diag_process}) and then describe the architecture of each module (Section~\ref{sec:pathfound}).

\subsection{An Agentic Diagnostic Process: Exploration, Execution, and Exploitation}
\label{sec:diag_process}
The diagnostic process of PathFound operates as a dynamic loop involving the three proposed modules following a 3E fashion: exploration, execution, and exploitation, as shown in Fig.~\ref{fig:overview}. 
Notably, these three stages need not be executed in sequence. Besides the typical path we investigate, a direct exploration-exploitation diagnosis and a much longer exploration-execution-exploration-execution-exploitation diagnosis are also compatible with our agentic system, representing a confident diagnosis in the early stage and a hesitant diagnosis after the first round of evidence acquisition, respectively. The difference lies in the diagnostic reasoner's prompts, detailed in~\ref{app:prompt}, and we provide cases to reflect complex routes in~\ref{app:case}. 

\paragraph{Exploration: The Initial Diagnosis Stage} 
In an earlier stage of diagnosis, only a little information is available, such as the patients' clinical history and corresponding H\&E slides. This may not support a precise conclusion. Therefore, rather than reaching a premature conclusion, it is preferred to generate a \textbf{differential diagnosis}, a list of potential diseases ranked by likelihood. More crucially, it is also encouraged to formulate an \textbf{action plan} that specifies which information should be collected to enable a more precise diagnosis. As for our model, things start with calling the \textit{slide highlighter} and the \textit{vision interpreter}. Without any a priori information, the \textit{slide highlighter} first initializes with the toolkit for generic pan-cancer abnormality to generate a probability map of common suspicious regions, and then samples representative RoIs likely to contain tumorous features using the diversity-aware region selection strategy. These RoIs are processed by the \textit{vision interpreter} to produce general visual findings of the slide. With this preliminary information retrieved, PathFound calls the \textit{diagnostic reasoner} to formulate an initial differential diagnosis that guides possible diagnostic directions, as well as an action plan identifying the following steps to refine the diagnosis, including re-observing the slide with specific purposes and ordering further examinations. 

\paragraph{Execution: The Evidence-Seeking Stage} Guided by the plan, PathFound actively acquires further evidence via two channels: internal re-inspection of slides with a specific purpose and external requests for further laboratory evidence. The re-inspection starts with re-triggering the \textit{slide highlighter}. At this time, the \textit{slide highlighter} is equipped with a different toolkit tailored to the requested visual evidence. Depending on the target granularity, a diversity-aware region selection strategy or a fine-grained entity selection strategy is adopted to propose relevant RoIs. 
The \textit{vision interpreter} operates in a more controlled manner to produce more focused visual judgments, taking into account the specific task. 
The external requests, however, rely on pathologists' further input, which fosters a multi-turn interaction. In this work, we leverage an LLM to generate these examination results following RAGES proposed by the work~\citep{anonymous2025reinforcement}. 

\paragraph{Exploitation: The Final Decision Stage} Upon receiving the supplementary evidence requested, PathFound transitions into an exploitative reasoning phase. The \textit{diagnostic reasoner} is encouraged to provide a precise diagnosis by integrating previous diagnostic context with newly acquired findings to resolve ambiguities. This consolidation includes validating or discarding earlier hypotheses in light of detected visual entities and laboratory results, and refining the diagnosis based on specific visual findings.

\subsection{PathFound}
\label{sec:pathfound}

\subsubsection{Slide Highlighter}

\begin{figure}[ht]
  \centering
  \includegraphics[width=\linewidth]{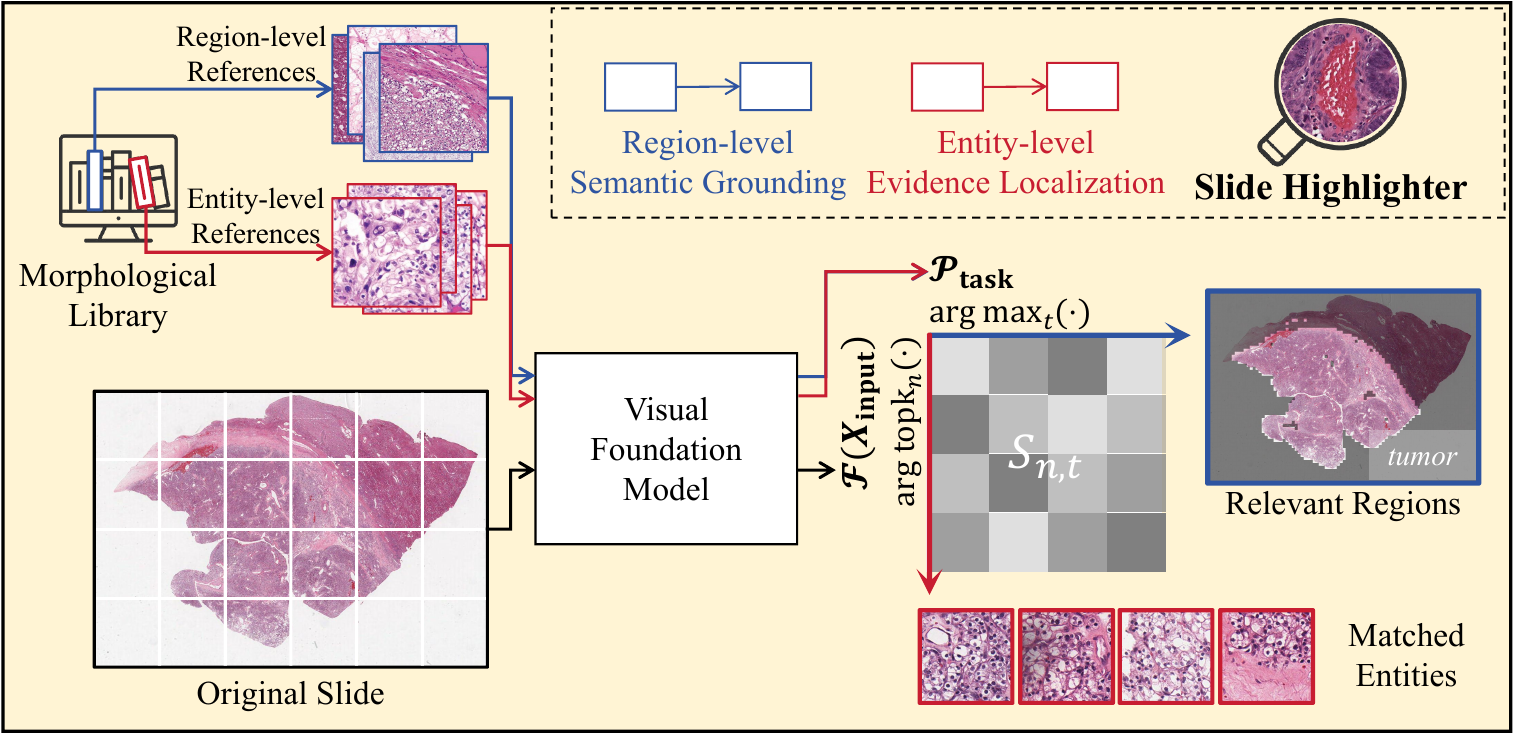}
  \caption {A detailed view of the slide highlighter. }
  \label{fig:detail}
\end{figure}

WSIs, with their gigapixel resolution, challenge standard vision-language models. To bridge this gap, the slide highlighter (Fig.~\ref{fig:detail}) summarizes a WSI into a compact set of representative RoIs tailored to different diagnostic goals. Drawing from few-shot paradigms~\cite{wang2019simpleshot}, we exploit existing VFMs in pathology to perform semantic retrieval. The highlighting pipeline comprises three stages: morphological library construction, task specific toolkit building, and region highlighting (region-level semantic grounding or entity-level evidence localization).

\paragraph{Morphological Library Construction} To enable flexible highlighting across diverse diagnostic tasks, we construct a library of toolkits containing different morphological prototypes. This library encodes visual representations of various pathological entities, ranging from coarse-grained categories (e.g., pan-cancer tumor vs. non-tumor) to fine-grained findings (e.g., specific nuclear grades). 

Formally, to construct the library, we define a set of $L$ morphological descriptions $\mathcal{T} = \{\tau_1, \tau_2, \dots, \tau_L\}$. For each description $\tau_l$, we collect a support set with $M$ reference images $X^{\tau_l}_{\text{ref}} = \{x^{\tau_l}_{\text{ref},i}\}_{i=1}^{M}$.  A frozen VFM $\mathcal{F}$ is then utilized to map these images into the latent space and the prototype feature $p_{l}$ is computed by averaging these embeddings: $p_{l} = \frac{1}{M} \sum_{i=1}^{M} \mathcal{F}(x^{\tau_l}_{\text{ref},i})$. The collection of all prototype features $\mathcal{P}=\{p_{l}\}$ serves as the semantic anchors for the subsequent highlighting process. 

\paragraph{Task-Specific Toolkit Building}
In practice, different diagnostic tasks focus only on a subset of morphological prototypes from the global library, covering various aspects (e.g., tumor vs. non-tumor for tumorous region detection). We therefore construct a task-specific toolkit $\mathcal{P}_{task} = \{p_{1}, p_{2},...,p_{T}\} \subset \mathcal{P}$, together with its associated description set $\mathcal{T}_{task} = \{\tau_1, \tau_2, \dots, \tau_T\}  \subset \mathcal{T}$, by selecting task-relevant prototypes from the morphological library. 

The task-specific toolkit is used to guide slide-level analysis. An input WSI is cropped into $N$ patches $X_{input} = \{x_{1}, x_{2}, ...,x_{N}\}$ at the same magnification level used during prototype construction. Each patch is embedded by $\mathcal{F}$ and compared against all prototypes in $\mathcal{P}_{\text{task}}$ using cosine similarity, yielding a similarity matrix $\mathbf{S} \in \mathbb{R}^{N \times T}$, where $S_{n,t} = \text{cos}(\mathcal{F}(x_{n}), p_{t}) $. 

Depending on the diagnostic objective, $\mathbf{S}$ can be exploited in two complementary task formulations: \textit{region-level semantic grounding}, which assigns each patch to its most relevant morphological description for slide understanding, and \textit{entity-level evidence localization}, which searches the entire slide for regions most strongly associated with a specific prototype. 

\paragraph{Region-level Semantic Grounding}
To capture relevant regions for a given task, we assign prototypes to each patch $x_n$ based on the similarity matrix $\mathbf{S}$: 
\begin{equation}
    t_n = \arg\max_{t} S_{n,t},\quad
    \tau^{*}(x_n) = \tau_{t_n}.
\end{equation}
Each patch $x_n$ is assigned to the description $\tau^{*}(x_n)$ with the highest similarity score (e.g., tumor regions for a specific cancer type). 
For a specific task, we then define the semantic concept to be highlighted (e.g., tumor regions), which is represented by $\mathcal{T}_{\text{Highlight}} \subset \mathcal{T}_{\text{task}} $. 
Then, the corresponding highlighted regions are $H_{R} = \{x_{n} \mid \tau^{*}(x_n) \in \mathcal{T}_{Highlight}\}$. To reduce the computational cost, we apply top-$k$ sampling and stochastic sampling on $H_{R}$ to ensure both high relevance and diversity. 

\paragraph{Entity-level Evidence Localization}
When the goal is to localize targeted pathological entity, we focus on finding patches that match the prototype most closely across the WSI. Therefore, for each entity description $\tau_{t}$, we select the tok-$k$ most similar parches, and the selected entity regions are defined as $H_{E} = \{ x_{n} \mid n \in \operatorname*{arg\,topk}_n (S_{n,t}) \}$, where $\operatorname*{arg\,topk}_n (\cdot)$ returns the indices of $k$ items with the highest scores.

\subsubsection{Vision Interpreter}
Upon distilling the representative RoIs, the vision interpreter functions as a semantic bridge, translating high-dimensional visual features into coherent medical descriptions. We implement this using a VLM comprising a visual encoder for feature extraction, a modality adapter for alignment, and a large language model (LLM) decoder for text generation. This design enables the system to articulate visual findings in natural language, facilitating subsequent reasoning.

\begin{table*}[ht]
\caption{Data for pathological domain adaptation of the vision interpreter.}
\centering
\label{tab:vlmdata}
\begin{tabular}{p{4cm}p{5cm}lp{4.5cm}}
\hline
Training Data&Training Task& Size & Data Source \\ \hline

&&229,874& PathGen \\

&&223,152& PathCap\\

&&205,506& Quilt-1M\\

&& 12,325& Public Websites\\

&& 179,143& Textbooks\\

\multirow{-6}{4cm}{Visual-linguistic Alignment \newline (885,521)}&\multirow{-6}{*}{Image-caption Alignment}&35,521& TCGA\\ \hline

% \multirow{-7}{*}{Visual-linguistic Alignment}& Similarity Assessment & 100,000 & TCGA \\\hline
&&890& TCGA Pan-cancer\\

&&741& TCGA-NSCLC\\

&\multirow{-3}{*}{Slide-level Diagnosis}&829&TCGA-RCC\\ \cline{2-4}

&Patch-level Subtyping&97,906&Public Task Datasets\\\cline{2-4}

&Multiple Choice&6,668&Private Question Bank\\\cline{2-4}

&&200,480& PathGen-Instruct\\

&&62,440& Quilt-Instruct\\

&\multirow{-3}{*}{Conversation}& 10,134 &TCGA \\ \cline{2-4}

& General Multimodal Instructions & 690,045 &LLaVA-NeXT-780k\\ \cline{2-4}

\multirow{-10}{4cm}{Instruction Tuning \newline (1,286,355)}& Text-only Medical Instructions &216,222 & Private \\
\hline
\end{tabular}
\end{table*}

\paragraph{Pathological Domain Adaptation} Generic VLMs often lack the specialized knowledge required for histopathology. To bridge this gap, we collect pathology-specific data to adapt the model to the pathological domain. Our training data can be classified into two parts. 
The \textit{visual-linguistic alignment} data aligns the visual latent space with medical terminology by exposing the model to large-scale pathological image-text pairs, enabling it to recognize and name fundamental pathological features. We collected 885K image-caption pairs sourced from public datasets~\citep{pathgen, zhang2020evaluating, ikezogwo2023quilt}, websites\footnote{https://www.webpathology.com} and textbooks. 
The \textit{instruction tuning} data facilitates transition from simple recognition to complex task solving. This data equips the model to handle varied downstream applications, effectively teaching it to follow diagnostic commands and structure its responses in accordance with clinical norms. The instruction tuning data consists of three parts. The pathological-specific multimodal data is sourced from various datasets, including TCGA\footnote{https://portal.gdc.cancer.gov}, public task datasets~\citep{breakhis,chaoyang, crc100k,lc25k,panda,gamper2020pannuke,PatchGrastricADC2022,pcam,sicapv2} and instruction datasets~\citep{pathgen, quilt-llava}, as well as private data such as TCGA annotations and a question bank. Data samples from public sources are cleaned and restructured with Gemini-2.5-Pro to ensure quality. 
Meanwhile, to preserve the original multimodal capability and medical understanding, we also incorporate a general multimodal instruction dataset, LLaVA-NeXT-780k~\citep{llavanext}, and private text-only medical data. The detailed data summary is listed in Table~\ref{tab:vlmdata}.

\paragraph{Visual In-Context Learning} While the fine-tuned model captures general knowledge, specific diagnostic tasks often require fine-grained comparison or adaptation to rare cases. To address this, we incorporate an in-context learning (ICL) mechanism in particular scenarios. 
Instead of relying solely on the model's textual understanding, we use the reference images $X_{\text{ref}}^{\tau}$ in the toolkit library of description $\tau$ as visual contexts. By concatenating these reference examples with the query RoI, we explicitly condition the VLM's generation on the provided context and purpose. This approach encourages the model to perform analogical reasoning by mapping the visual relationships observed in the references to the query image. This not only improves trustworthiness by grounding the output in concrete examples but also enhances the model's ability to generalize to unseen patterns through inductive inference. 

\subsubsection{Diagnostic Reasoner}
The diagnostic reasoner acts as the central decision engine of PathFound. It is responsible for orchestrating the entire agentic process, synthesizing multimodal evidence (clinical history, visual findings, and appended testing results), and managing user interactions. 
To endow the model with the capability for complex clinical reasoning with agents, we adopt an RLVR framework, i.e., the Group Relative Policy Optimization (GRPO) algorithm~\citep{deepseekai2025deepseekr1incentivizingreasoningcapability}. To adapt to the possibility-ranking nature of ambiguous diagnoses, we use the diagnostic reward in the work~\citep{anonymous2025reinforcement}. The original reward contains three parts: a format reward, a rank-sensitive diagnostic reward $R_d$, and an examination consistency reward $R_e$. To accommodate re-trigger of tools, we further implement a tool-call reward $R_t$. The whole reward can be formulated as follows.

\begin{equation}
\label{eq:reward_all}
R(O,Y) = 
\begin{cases} 
\begin{aligned}
& R_d(\sD,Y) + R_e(\sT,\sD)  \\
& \;\;+ R_t(\sC, \sD) - P_h \times \bm{1}_{\mathrm{c}}
\end{aligned}   & n_f = 0, \\
-P_f \times n_f & n_f \neq 0,
\end{cases}
\end{equation}
where $O$ is the model output, and $Y$ the ground truth diagnosis. $n_f \in \{0,1,2\}$ denotes the number of format errors, including missing think/answer pairs or improperly presented answers. $\sD$ represents the extracted ordered diagnosis list from $X$, $\sT$ the test list, and $\sC$ the tool calling list. $P_f$ and $P_h$ denote the format and the hacking penalty, respectively. $\bm{1}_{\mathrm{c}}$ indicates the occurrence of hacking behavior like conflicting diagnoses and overly vague diagnoses discussed in~\citet{fuhrman_isup}. 

The rank-sensitive diagnostic reward $R_d$ and examination consistency reward $R_e$, assisted by an LLM judge, are formulated as,

\begin{equation}
\label{def:d_reward}
R_d(\sD,Y) = 
\begin{cases} 
\frac{\exp(-i/\alpha)}{\sum_{j=1}^{|\sD|}\exp(-j/\alpha)} & Y=D_i \in \sD,\\
0 & Y \notin \sD,
\end{cases}
\end{equation}

\begin{equation}
R_e(\sT,\sD) = 
\begin{cases} 
B_e & \text{$\sT$ can differentiate $\sD$ effectively}, \\
0 & \text{$\sT$ have no severe conflicts with $\sD$}, \\
-B_e & \text{$\sT$ contain some severe problems},
\end{cases}
\end{equation}
where $\alpha>0$ is a hyperparameter controlling the score distribution and $B_e$ is the value of consistency bonus. The judge LLM is used to determine whether and at which position the proposed diagnoses align with the ground truth in $R_d$, as well as whether the proposed examinations are proper.

For the tool calling reward $R_t$, we use predefined rules to determine whether it should receive a bonus. This is due to two reasons. First, the current tool set is limited in scale, so we can write rules to complete the judging work. And more importantly, the incorporation of a third LLM judge would significantly increase the whole training time. 

\begin{equation}
R_t(\sC,\sD) = 
\begin{cases} 
B_t & \text{$\sC$ are properly called w.r.t. $\sD$}, \\
0 & \text{No tools should be called w.r.t. $\sD$}, \\
-B_t & \text{$\sC$ have falsely called tools w.r.t. $\sD$},
\end{cases}
\end{equation}
where $B_t$ is the value of the tool calling bonus.

\section{Experiments}
\subsection{Implementation}
\label{sec:im}

\paragraph{Slide Highlighter Setup}

We employ the foundation model UNI-2 as the feature extractor. The morphological library is constructed from public datasets, including the TCGA dataset and the AGGC 2022 challenge~\citep{huo2024comprehensive}. The toolkit building details can be found in~\ref{app:imple}. The top-3 most similar RoIs across all tumor prototypes at both 10× and 20× magnifications, augmented by two other randomly selected tumorous patches at 10× magnification, yielding a total of 8 RoIs, are selected for pan-cancer screening. For detailed RCC subtyping, the two randomly selected RoIs are replaced with RoIs similar to the prototypes augmented with $K$-means to strengthen the relevance. For nuclear grading, the top 5 most similar patches at 20× magnification of each category are fed into the vision interpreter with ICL. For Gleason grading, the 20× highlighted map, together with the measured areas, is used to determine the Gleason score. For invasion detection, the top 5 most similar patches at 5× and 10× magnifications are selected for the following ICL judgment.

\paragraph{Vision Interpreter Configuration}
We utilize Qwen2.5-VL-7B-Instruct~\citep{qwen2.5-vl} as the backbone VLM and leverage the LLaMA-Factory framework~\citep{zheng2024llamafactory} to fine-tune the model. The training data consists of two distinct compositions, as detailed in Table~\ref{tab:vlmdata}. 
We perform full fine-tuning for three epochs using the AdamW optimizer, with the visual encoder, the modality adapter and the LLM all kept unfrozen. The learning rate is set to $2 \times 10^{-6}$ with a cosine decay scheduler and a warm-up ratio of 0.01. The batch size is 32. Training takes approximately 30 hours on 8 NVIDIA H200 GPUs. When using ICL, a total of 10 reference images are selected to aid judgment.

\paragraph{Diagnostic Reasoner Development} We adopt Qwen2.5-32B-Instruct~\citep{qwen2.5} as the base model for the diagnostic reasoner. The development involves a rigorous data construction pipeline followed by RLVR. 
We sourced 1,026 high-quality raw cases from the \textit{Daka}\footnote{https://www.dakapath.com/} online repository (373 cases) and the \textit{Chinese Journal of Pathology} (653 cases). We then employed GPT-4 to extract structured clinical information and ground truth diagnoses. We then collect 118 cases of pure text from the early cases in Xijing Hospital, dated before 2022. These cases are about renal cell carcinoma, and we assigned ground truth tool calling list to them following the rules: (1) for specific subtype, the tools should include corresponding renal cell observing tool; (2) for clear cell and papillary cell carcinoma, the tools should also include a nuclear grading tool; (3) once the microscopic description involves invasions, the tools should include an invasion detecting tool. To simulate the information discrepancy, we split the cases with further testing results into two versions: one without the additional results, corresponding to a lower $\alpha_1=0.5$, and the other with these results, using $\alpha_2=2$. The final training corpus includes 1529 data samples, containing 239 cases of the tool calling list (only 3 of them calling the invasion tools). 947 cases contain only preliminary information, and the remaining 582 have further examination results. 
For the reward function, $P_f$, $P_h$, $B_e$, and $B_t$ are set to 0.5, 0.3, 0.1, and 0.1, respectively. The training is implemented using OpenRLHF~\citep{hu2024openrlhf} with vLLM~\citep{vllm} for rollouts and conducts on 16 H200 GPUs for RL and 8 H100 GPUs for LLM-based judging. The training completes in about 120 hours. The training batch size is 32 with a micro batch 2. The rollout batch size is 64 with a micro of 4. For each data item, 4 samples are generated, with a maximum prompt and response lengths 2048. The actor learning rate is $5\times 10^{-7}$ with a total of 20 episodes. The rest hyper-parameters are by default.

\subsection{Evaluation Configurations} 
\subsubsection{Clinical Scenes}
\paragraph{Scene 1: Renal Cell Carcinoma (RCC)} 
RCC is an ideal testbed for evidence-seeking diagnostic reasoning because of its hierarchical, modular diagnostic workflow. It can be categorized into various subtypes, among which the following three are the most common: clear cell RCC (ccRCC), papillary RCC (pRCC), and chromophobe RCC (chRCC)\footnote{In TCGA, the ccRCC is referred to as the KIRC group, pRCC as KIRP, and chRCC as KICH.}. Accurate subtyping requires recognizing subtype-specific cytological morphologies and architectural patterns, and in borderline cases, interpreting protein expression profiles from further examinations. 
For confirmed ccRCC and pRCC cases, nuclear grading is critical for risk stratification and prognostic assessment. This grading process naturally demands a ``look-back'' mechanism that requires attention to fine-grained features such as nucleolar prominence, irregular nuclear contours, and cellular anaplasia. Accordingly, we abstract the evaluation into a two-step challenge: primary RCC subtyping followed by nuclear grading.

We curate an evaluation dataset sourced from TCGA-RCC, consisting of 102 ccRCC, 59 pRCC, and 25 chRCC samples. Given the limited availability of reliable grading annotations, we further identify a subset of 8 cases for a pilot assessment of nuclear grading. To further test model robustness and out-of-distribution (OOD) generalization, we incorporate additional cases from Xijing Hospital (Xijing) dated after 2024 and Fudan University Shanghai Cancer Center (SCC). Xijing provides 53/1/4 ccRCC/pRCC/chRCC cases, with 54 cases offering ground-truth nuclear grades. SCC includes 37/27/37 ccRCC/pRCC/chRCC cases, and 62 of these provide nuclear grading annotations.

\paragraph{Scene 2: Prostatic Carcinoma} 
Prostatic carcinoma represents another clinically significant and morphologically diverse malignancy. The TCGA-PRAD group focuses on adenocarcinoma of various types, including prostatic acinar adenocarcinoma and prostatic ductal adenocarcinoma. The diagnostic workflow centers on characterizing glandular architectural distortions and cytological atypia, and its prognostic assessment relies heavily on the Gleason grading system. Gleason grading is inherently hierarchical because the model must first identify tumorous regions, then analyze their severity, and finally generate the primary and secondary Gleason scores for the case, making this an ideal test of multi-step diagnostic reasoning. Notably, the diagnostic reasoner is not explicitly trained to invoke prostate-related tools, allowing PRAD to serve as a natural OOD challenge for evaluating tool-calling generalization. We select 447 cases with complete and reliable Gleason score records from TCGA-PRAD to evaluate both the diagnosis and detailed grading performance of PRAD.

\paragraph{Scene 3: Lymphovascular and Perineural Invasion}
Lymphovascular invasion (LVI) and perineural invasion (PNI) are critical prognostic factors across numerous cancer types, reflecting the presence of tumor cells within lymphatic or vascular channels and nerves, and signaling an increased risk of metastatic progression. Invasion detection requires recognizing subtle morphological cues, such as tubular structures and intraluminal tumor clusters, making it a fine-grained yet clinically indispensable diagnostic task. Moreover, because LVI and PNI manifest across a broad spectrum of malignancies, they provide a natural and rigorous setting for evaluating pan-cancer diagnostic generalization. From TCGA, we curate 207 invasion-related cases, comprising 102 positive (LVI: 90, PNI: 12) and 105 negative samples (LVI: 91, PNI: 14) across 28 tumor types. We refer to this benchmark as the 2-class TCGA-Invasion dataset, designed to systematically assess a model’s generalization capacity and its evidence-seeking behavior in pan-cancer invasion detection.

\begin{table*}[htb]
\centering
\caption{Overall diagnostic performance on five datasets covering three scenes. *: The model cannot correctly respond to an evidence-seeking prompt. **: The model is not involved in an in-house evaluation considering data privacy. \textbf{Bold} suggests the best performance and \textit{italic} suggests the second.}
\label{tab:diagnostic}
\begin{tabular}{lcccccccccc}
\hline
\multirow{2}{*}{Model} & \multicolumn{2}{c}{TCGA-RCC}    & \multicolumn{2}{c}{Xijing-RCC}  & \multicolumn{2}{c}{SCC-RCC}     & \multicolumn{2}{c}{TCGA-PRAD}   & \multicolumn{2}{c}{TCGA-Invasion} \\ \cmidrule(lr){2-3} \cmidrule(lr){4-5} \cmidrule(lr){6-7} \cmidrule(lr){8-9} \cmidrule(lr){10-11}
                       & OP             & ES             & OP             & ES             & OP             & ES             & OP             & ES             & OP              & ES              \\ \hline
Qwen3-VL               & 34.10          & 44.85          & \textit{31.90} & 39.98          & 36.33          & 43.30          & 78.08          & 86.30          & 59.42           & 72.46           \\
InternVL3.5            & 33.33          & 54.82          & 26.44          & \textit{64.94} & \textit{39.14} & \textit{57.59} & 78.99          & 89.26          & 62.32           & 77.29           \\
PathGen-LLaVA          & 29.18          & *              & 20.40          & *              & 28.79          & *              & 52.05          & *              & 41.55           & *               \\ \hline
Gemini-2.5             & 31.04          & 66.59          & **             & **             & **             & **             & \textit{81.27} & 90.60          & \textit{67.15}  & \textit{80.19}  \\
GPT-5                  & \textit{38.34} & \textit{71.52} & **             & **             & **             & **             & 68.90          & \textbf{92.47} & 61.84           & 79.23           \\ \hline
Ours                   & \textbf{59.24} & \textbf{92.28} & \textbf{71.84} & \textbf{97.13} & \textbf{52.29} & \textbf{91.79} & \textbf{84.56} & \textit{92.17} & \textbf{69.57}  & \textbf{81.64}  \\ \hline
\end{tabular}
\end{table*}

\subsubsection{Evaluation Protocols and Metrics.}
We design two distinct evaluation protocols to benchmark the models' performance on diagnosis: 
\paragraph{One-Pass Inference Protocol (OP)} Models are provided with the slide and basic patient metadata and are prompted to output a final diagnosis immediately. This measures the baseline capability of passive recognition and mimics the traditional usage. 
\paragraph{Evidence-Seeking Protocol (ES)} This protocol mimics the clinical workflow. Models are required to first provide a \textit{differential diagnosis} and formulate \textit{requests for further evidence}. After receiving the requested information (visual or clinical), they output the \textit{precise diagnosis}. To simulate examination results during evaluation, we use RAGES to generate results from Gemini-2.5-Pro. 

Given the class imbalance in the RCC test set, we report \textbf{Balanced Accuracy} (BAcc) to prevent bias towards the prevalent ccRCC class. For TCGA-PRAD and TCGA-Invasion, we report the \textbf{Accuracy} (Acc).

For the \textit{nuclear grading} task, we first present a qualitative comparison of model predictions against ground truth in TCGA. Then, we report the quantitative performance of models on datasets from Xijing and SCC. To note, given the differences and relationships between the Fuhrman (TCGA) and ISUP (Xijing and SCC) grading systems, we report the accuracy in differentiating between low (grades 1 and 2) and high grades (grades 3 and 4), which has shown high consistency across both grading systems \cite{fuhrman_isup}. 
Inspired by the tool-call success rate, we also report the \textbf{proactive evidence mentioning rate (PEMR)}, i.e., the frequency with which a model explicitly mentions nuclear grading criteria during reasoning, to reflect models' awareness of incorporating key observation points. For the \textit{Gleason grading} task, we evaluate two levels of performance: the primary pattern accuracy (like Gleason Grade 3) and the combined score of both primary and secondary patterns (like Gleason score 3+4). The grade involves Gleason patterns 3 to 5, resulting in 3-class and 9-class tasks. PEMR is also evaluated for Gleason grading. For \textit{invasion detection}, we report precision, recall, F1 score, and PEMR. 

\subsubsection{Baselines}
We compare PathFound against a comprehensive set of multimodal models, categorized into two groups. We select open-source models with parameter scales comparable to ours, including Qwen3-VL-32B-Instruct~\citep{qwen3vl} (Qwen3-VL), InternVL3.5-38B~\citep{internvl35} (InternVL3.5), and the pathology-optimized PathGen-LLaVA-13B~\citep{pathgen} (PathGen-LLaVA). We also benchmark against closed-source frontier models under open-source datasets, including Gemini-2.5-Pro (Gemini-2.5) and GPT-5-High (GPT-5), to assess the upper bound of performance.

\section{Results Analysis}

\subsection{Scene 1: Renal Cell Carcinoma}

\paragraph{Diagnostic Performance on RCC Subtyping}

Table~\ref{tab:diagnostic} benchmarks the performance of PathFound against state-of-the-art multimodal models on the RCC subtyping task across three data sources. We report the balanced accuracy under two inference protocols, as previously clarified. PathGen-LLaVA is evaluated solely on the one-pass protocol because it cannot follow a multi-turn instruction fashion. Gemini-2.5 and GPT-5 only participate in the TCGA cohort to protect the privacy of in-house data. 
Across all models, the evidence-seeking protocol yields substantial performance gains compared to the one-shot approach. This empirically validates our core hypothesis that mimicking the clinical hypothesis-driven workflow is superior to read-once predictions. More interestingly, frontier proprietary models exhibit a dramatic improvement (over 30\%) when switched to the evidence-seeking mode, whereas smaller open-source models show moderate gains (10-20\%). This suggests that larger models possess stronger \textit{instruction-following} and \textit{reasoning} capabilities, allowing them to utilize better the supplementary evidence provided in the second turn. PathFound achieves state-of-the-art performance in both settings, with a remarkable final balanced accuracy of 92.28\%. Models participating in the OOD evaluation demonstrate high robustness under cross-site conditions, which can be attributed to the intrinsic generalization capability of large models. 

\begin{figure}[ht]
  \centering
  \includegraphics[width=\linewidth]{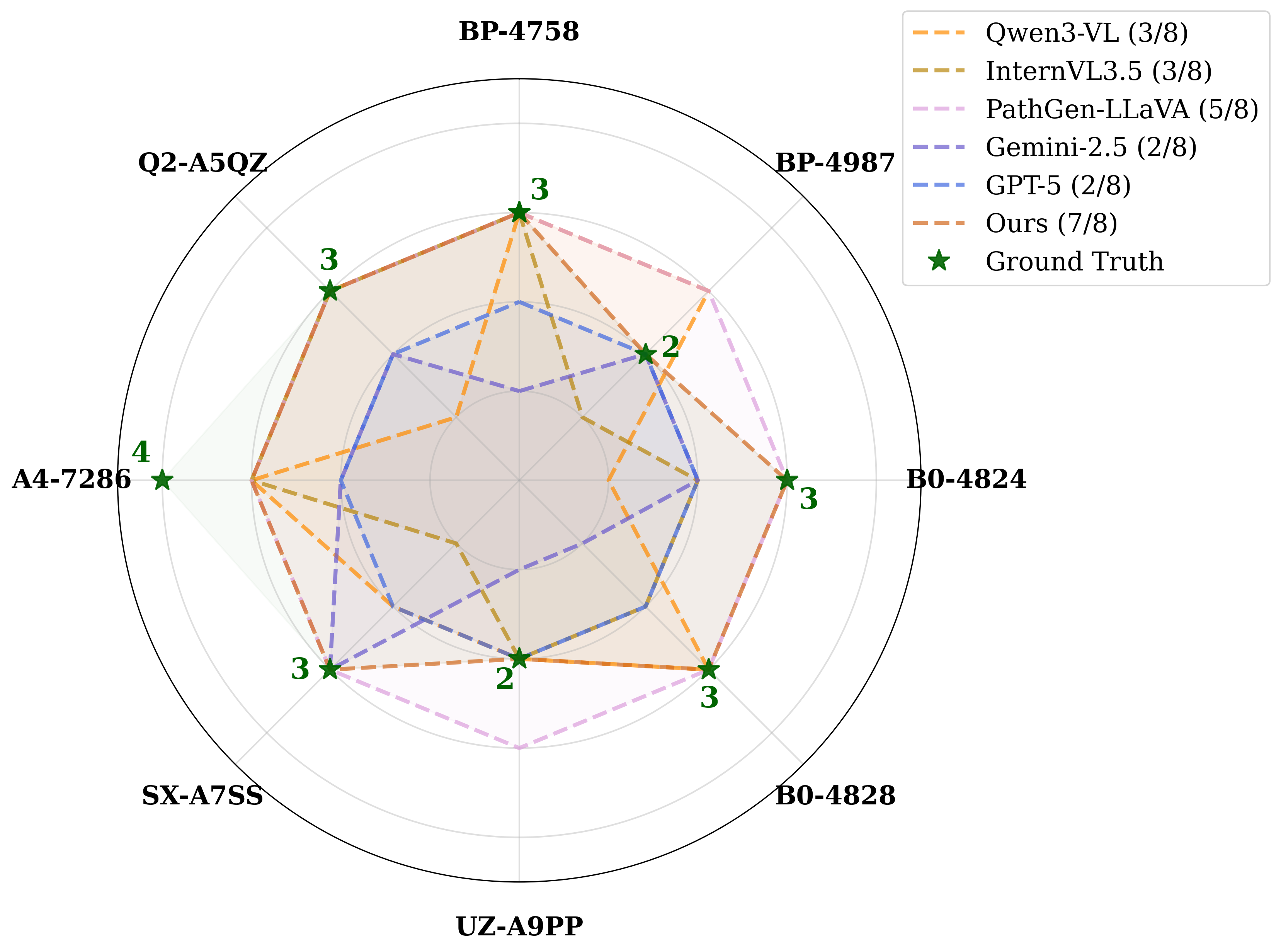}
  \caption {Qualitative Results on Nuclear Grading from TCGA-RCC.}
  \label{fig:grade}
\end{figure}

\begin{table*}[ht]
\centering
\caption{Overall performance on four datasets covering three fine-grained tasks. NG: nuclear grading. P.Acc: primary accuracy. C.Acc: combined accuracy. Prec.: precision. Rec.: Recall. \textbf{Bold} suggests the best performance and \textit{italic} suggests the second.}
\label{tab:fine_task}
\begin{tabular}{lp{0.70cm}<{\centering}p{0.82cm}<{\centering}p{0.70cm}<{\centering}p{0.82cm}<{\centering}p{0.82cm}<{\centering}p{0.82cm}<{\centering}p{0.82cm}<{\centering}p{0.75cm}<{\centering}p{0.75cm}<{\centering}p{0.75cm}<{\centering}p{0.82cm}<{\centering}}
\hline
\multirow{2}{*}{Model} & \multicolumn{2}{c}{Xijing-NG}   & \multicolumn{2}{c}{SCC-NG}      & \multicolumn{3}{c}{Gleason Grading}              & \multicolumn{4}{c}{Invasion Detection}                            \\
     \cmidrule(lr){2-3} \cmidrule(lr){4-5} \cmidrule(lr){6-8} \cmidrule(lr){9-12}
     % \hline
   & Acc            & PEMR           & Acc            & PEMR           & P.Acc         & C.Acc         & PEMR           & Prec.      & Rec.         & F1             & PEMR           \\ 
   \hline
Qwen3-VL               & 22.22          & \textit{4.76}  & 46.77          & \textit{6.93}  & 48.40          & 23.06          & 12.53          & 44.00          & 24.44          & 31.42          & 1.93           \\
InternVL3.5            & \textit{55.56} & 0.00           & \textit{51.61} & 2.97           & 50.23          & 22.83          & 7.61           & \textit{46.67} & 62.22          & 53.33          & 2.89           \\
PathGen-LLaVA          & 24.07          & 3.17           & 20.97          & 0.00           & \textit{51.60} & 17.12          & 16.78          & 44.87          & 38.89          & 41.67          & 0.97           \\ \hline
Gemini-2.5             & **             & **             & **             & **             & 51.14          & 22.83          & 18.34          & 41.75          & \textbf{90.00} & \textit{57.04} & \textit{7.73}  \\
GPT-5                  & **             & **             & **             & **             & 49.32          & \textit{25.57} & \textit{20.36} & 40.81          & \textit{78.89} & 53.79          & \textbf{12.08} \\ \hline
Ours                   & \textbf{64.81} & \textbf{96.83} & \textbf{69.35} & \textbf{95.05} & \textbf{63.53} & \textbf{40.72} & \textbf{79.64} & \textbf{67.65} & 75.82          & \textbf{71.50} & 4.34           \\ \hline
\end{tabular}
\end{table*}

\paragraph{Study on Nuclear Grading}
Fig.~\ref{fig:grade} presents a qualitative assessment of fine-grained nuclear grading on a pilot subset of 8 cases from TCGA. Among all models, PathFound is the only model capable of proactively initiating nuclear grading without explicit user prompting. Other models require an additional question to attend to these subtle features, highlighting the advantage of our \textit{active evidence-seeking} mechanism. After being prompted, most baselines exhibit a systematic bias towards underestimation, i.e., predicting lower grades than ground truth. This suggests that VLMs may struggle to detect worst-case features (e.g., a few highly anaplastic cells) within a large slide. Among them, PathFound achieves the highest concordance with ground truth (7/8 correct), significantly outperforming random guessing.

A following quantitative experiment involving cases from Xijing and SCC is shown in Table~\ref{tab:fine_task}. In addition to accurate differentiation between low and high nuclear grades, we also report the PEMR for each model. Among all models, PathFound consistently reviews the slides for specific nuclear grading (PEMR > 95\% for both sites). It also delivers the best performance in nuclear grading. Yet, there remains substantial room for improvement in precise judgment.

\subsection{Scene 2: Prostate Adenocarcinoma}
 While \textit{Gleason Acc} assesses the ability to perform clinically relevant Gleason grading, including both primary pattern recognition and the combined Gleason score. We further report PEMR, reflecting whether the model actively recognizes the necessity of Gleason grading without explicit user prompting.

\paragraph{Diagnostic Performance on PRAD Detection} Table~\ref{tab:diagnostic} summarizes the diagnostic performance of different models on TCGA-PRAD. It evaluates whether the model can correctly identify the presence of prostate adenocarcinoma. Overall, most models achieve strong diagnostic accuracy and consistently benefit from the ES protocol. Compared with the OP setting, ES leads to notable performance gains across nearly all models, validating the effectiveness of iterative evidence collection. This improvement is particularly pronounced for proprietary large models, likely due to the abundant public availability of prostate pathology data and their superior reasoning capacity. After evidence-seeking, Gemini-2.5, GPT-5, and PathFound all exceed 90\% diagnostic accuracy. It is worth noting that GPT-5 exhibits relatively lower accuracy in the OP protocol. Manual inspection reveals that this is mainly caused by confusion between prostate adenocarcinoma and benign conditions such as prostatic hyperplasia, suggesting that subtle architectural cues may be overlooked in a single-pass inference.

\begin{table*}[ht]
\centering
\caption{Breakdown of the Diagnostic Trajectory. Metrics evaluate the quality of the initial hypothesis (Initial), the differential diagnosis list (DDx), and the final decision (Final). \textbf{Bold} suggests the best performance and \textit{italic} suggests the second.}
\label{tab:RCC_stage_bacc}
\begin{tabular}{lcccc}
\hline
Model & Initial BAcc & Initial DDx BAcc & DDx Length& Final BAcc \\ \hline
Qwen3-VL  & \textit{35.08} & \textit{92.61}     &  21.48    & 44.85     \\
InternVL3.5 & 29.17 & 83.05    &    19.04   & 54.82     \\ \hline
Gemini-2.5 & 18.92 & 62.01 &  8.12  & 66.59     \\
GPT-5  & 24.87 & 68.42      &  9.77   & \textit{71.52}     \\ \hline
% PathGen-LLaVA & - & - & - & -  \\
Ours  & \textbf{49.58} & \textbf{95.63} & 3.18 & \textbf{92.28}      \\ \hline
\end{tabular}
\end{table*}

\paragraph{Study on Gleason Grading}
Beyond the primary diagnosis, Gleason grading represents a more challenging fine-grained task that requires explicit assessment of glandular patterns. As shown in Table~\ref{tab:fine_task}, baseline models generally struggle with this task, particularly for predicting combined Gleason scores, indicating limited sensitivity to grading-relevant morphological patterns. In contrast, PathFound achieves the best performance in both primary Gleason grading and combined Gleason score prediction, with clear margins over competing methods. More importantly, PathFound demonstrates a substantially higher PEMR (79.64\%), proactively identifying the need for Gleason grading without additional user queries. This behavior closely mirrors real-world clinical practice, in which grading is integral to PRAD diagnosis rather than an optional follow-up.

\subsection{Scene 3: Invasion Detection}

\paragraph{Diagnostic Performance on TCGA-Invasion} 
Table~\ref{tab:diagnostic} reports the pan-cancer diagnostic accuracy on the TCGA-Invasion benchmark. As in previous scenes, switching from OP inference to ES inference consistently improves diagnostic accuracy across all models. PathFound achieves the best overall performance under both protocols, while proprietary models such as Gemini-2.5 and GPT-5 also demonstrate competitive results, reflecting their strong general diagnostic capabilities across diverse cancer types. 

\paragraph{Study on Invasion Detection}
As shown in Table~\ref{tab:fine_task}, different models exhibit markedly different precision-recall trade-offs. Gemini-2.5 achieves the highest recall (90.00\%), but this comes at the cost of substantially lower precision, indicating a tendency to over-predict invasive cases. GPT-5 shows a similar pattern, favoring recall over precision. In contrast, PathFound achieves the highest F1 score (71.50\%), outperforming all baselines by a clear margin of over 14\%. Considering that the evaluation dataset is more balanced than real-world clinical distributions, which typically contain a higher proportion of non-invasive cases, this balanced precision-recall profile is particularly desirable for practical deployment, where excessive false positives may lead to unnecessary downstream interventions. 
The overall PEMR remains relatively low across all models. Interestingly, proprietary models exhibit higher proactive mentioning rates, though still limited (maximum 12.08\%). This observation suggests that invasion assessment remains a cognitively demanding task for current multimodal models, often requiring expert intervention or targeted follow-up questions. It underscores the necessity of human-AI collaboration in scenarios where critical pathological features are easily overlooked.

\subsection{Ablation Study}

\subsubsection{Deconstructing the Diagnostic Trajectory on RCC Subtyping}

To understand how models arrive at the correct diagnosis in evidence-seeking mode, we dissect the process into three components: the accuracy of the initial hypothesis (Initial BAcc), the quality of the differential diagnosis list (DDx BAcc and Length), and the accuracy of the final decision (Final BAcc). The results are presented in Table~\ref{tab:RCC_stage_bacc}. 
Generic open-source VLMs often generate excessively long differential lists (approximately 20 candidates), thereby artificially inflating DDx accuracy. However, this approach reflects a lack of confidence rather than precise reasoning, as evidenced by their lower final accuracy. In contrast, PathFound maintains a concise differential list (average length 3.18) while achieving a respectable DDx accuracy of 95.63\%. Meanwhile, PathFound shows the highest accuracy across the initial hypothesis and the final decision. Even with a modest initial accuracy (49.58\%), it effectively leverages the evidence-seeking phase to correct its judgments, boosting the final accuracy to 92.28\%. When comparing Table~\ref{tab:diagnostic} and Table~\ref{tab:RCC_stage_bacc}, we observe an interesting phenomenon: most models yield less accurate primary diagnoses when allowed to provide a list of diseases, whereas the initial differential accuracy is more encouraging. This suggests a need for the following information to complete the diagnostic process and achieve precise outcomes.

\subsubsection{Impact of Evidence Sources}
We conduct an ablation study to quantify the contribution of different evidence sources: (1) visual re-observation (\textit{Further Look}) and (2) external examination results (\textit{Further Test}). As shown in Table~\ref{tab:ablation}, integrating either visual re-observation (+23.57\%) or external examination results (+28.91\%) improves performance over the baseline one-pass diagnosis. The inclusion of further test results yields a more significant boost. This mirrors clinical reality, where definitive diagnosis often relies on IHC or molecular tests when morphology alone is ambiguous. The combination of both sources achieves the peak accuracy of 92.28\%, confirming that our model effectively synthesizes multimodal information rather than relying on a single modality. 

\begin{table}[ht]
\centering
\caption{Ablation study of combining different further information.}
\label{tab:ablation}
\begin{tabular}{ccc}
\hline
Further Look      & Further Test      & TCGA-RCC BAcc   \\ \hline
    &   & 59.24 \\
\checkmark &   & 82.81 \\
 &  \checkmark & 88.15 \\
\checkmark & \checkmark & 92.28 \\ \hline
\end{tabular}
\end{table}

\subsubsection{Number of Input RoIs for The Vision Interpreter}

\begin{figure}[ht]
  \centering
  \includegraphics[width=\linewidth]{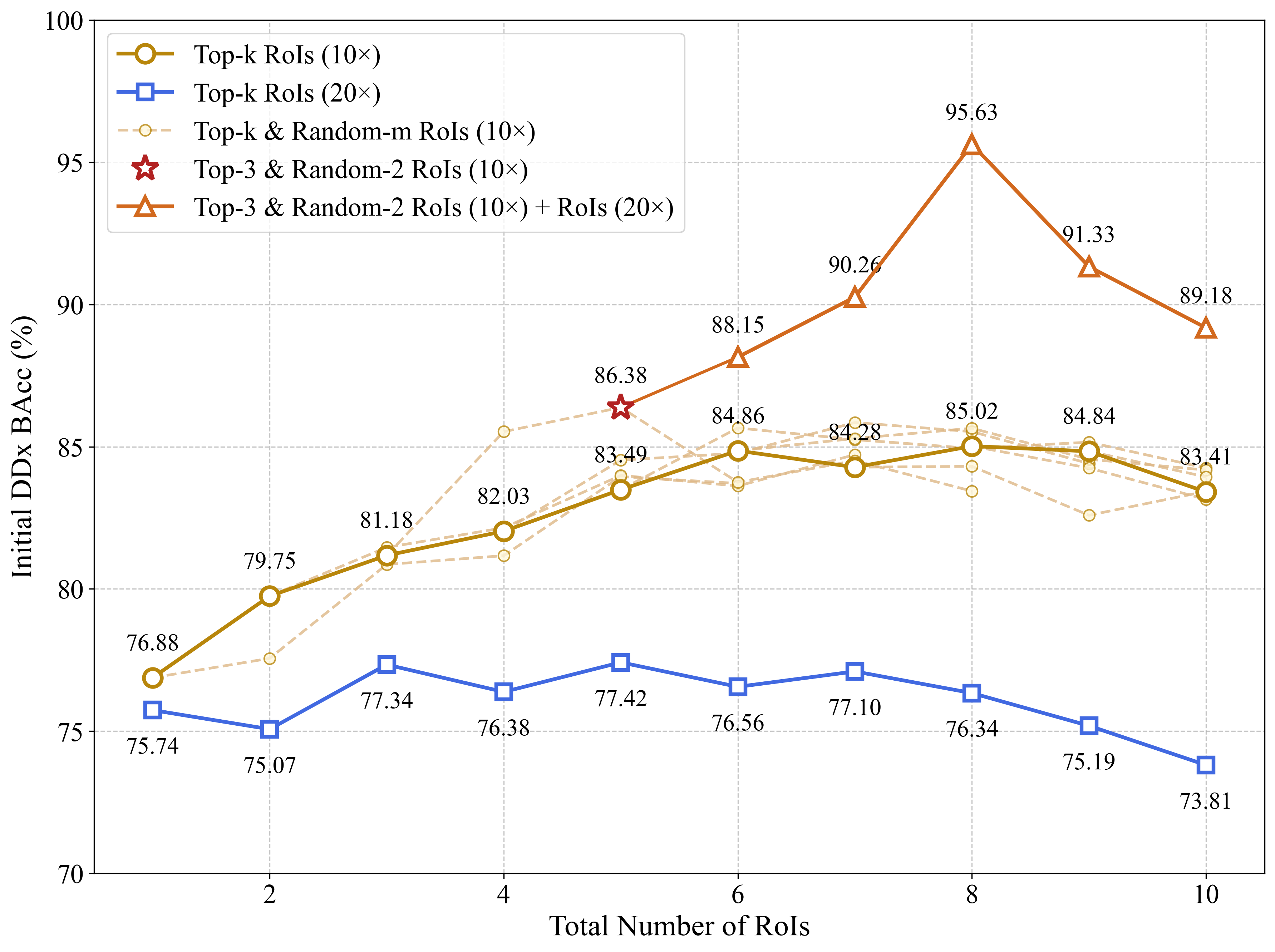}
  \caption {Ablation study on the combination of input RoIs when screening RCC.}
  \label{fig:num_roi}
\end{figure}

The selection of input RoIs plays a critical role in the agentic diagnostic process, as it directly affects contextual coverage, computational cost, and model performance. Key design factors include the number of RoIs, their magnifications, and the selection strategy. To identify a suitable configuration for pan-cancer screening, we conduct an ablation study on the TCGA-RCC subtyping task, using the initial differential diagnosis accuracy as the evaluation metric. The results are summarized in Fig.~\ref{fig:num_roi}. 

We first compare different numbers of top-$k$ similar RoIs extracted at 10$\times$ and 20$\times$ magnifications. RoIs at 10$\times$ consistently outperform those at 20$\times$. Increasing the number of RoIs is beneficial, but it gets saturated when six images are incorporated. Adding diversity-aware random images can improve results, but it requires balancing relevance (top-$k$ images) with exploration (random images). Considering both performance and context-length constraints, we adopt a configuration that combines the top-3 most similar RoIs with two randomly selected RoIs at 10$\times$ magnification as the basic setting. This setting achieves better performance than using only the top-$k$ similar RoIs. 
We further incrementally incorporate 20$\times$ RoIs. Performance peaks when the top-3 most similar 20$\times$ RoIs are included. Based on these observations, the final configuration for the initial diagnostic stage consists of three top similar RoIs at 10$\times$, two random RoIs at 10$\times$, and three top similar RoIs at 20$\times$. For the subsequent fine-grained RCC subtyping stage, replacing randomly selected RoIs with $K$-means based RoIs led to a 4.25\% absolute improvement in final accuracy, increasing from 88.03\% to 92.28\%. 

\begin{table}[ht]
\centering
\caption{Ablation study on the RoI selection plan of invasion detection. Each \checkmark suggests an inclusion of the top 5 most similar RoIs under certain magnification.}
\label{tab:inv_roi_num}
\begin{tabular}{cccccc}
\hline
5×  & 10× & 20× & Precision & Recall & F1    \\ \hline
\checkmark &            &            & \textbf{68.66}     & 69.61  & \textit{69.13} \\
           & \checkmark &            & 61.12     & 64.71  & 62.86 \\
           &            & \checkmark & 65.62     & 56.86  & 60.93 \\
\checkmark & \checkmark &            & \textit{67.65}     & \textbf{75.82}  & \textbf{71.50} \\
\checkmark &            & \checkmark & 67.39     & 63.72  & 65.50 \\
           & \checkmark & \checkmark & 60.04     & 59.80   & 59.92 \\
\checkmark & \checkmark & \checkmark & 66.67     & \textit{71.57}  & 69.03 \\ \hline
\end{tabular}
\end{table}

To account for variability in invasion size, we further perform ablation experiments on the magnifications used for selecting representative invasion-related RoIs. As shown in Table~\ref{tab:inv_roi_num}, the combination of 5$\times$ and 10$\times$ RoIs achieves the best overall balance between precision, recall, and computational efficiency. Notably, 5$\times$ RoIs provide coarse but comprehensive contextual information and are more likely to capture complete invasion regions, while incorporating additional RoIs substantially improves recall.

\subsubsection{The combination of ICL}
\label{sec:ab_icl}
\begin{figure}[ht]
  \centering
  \includegraphics[width=\linewidth]{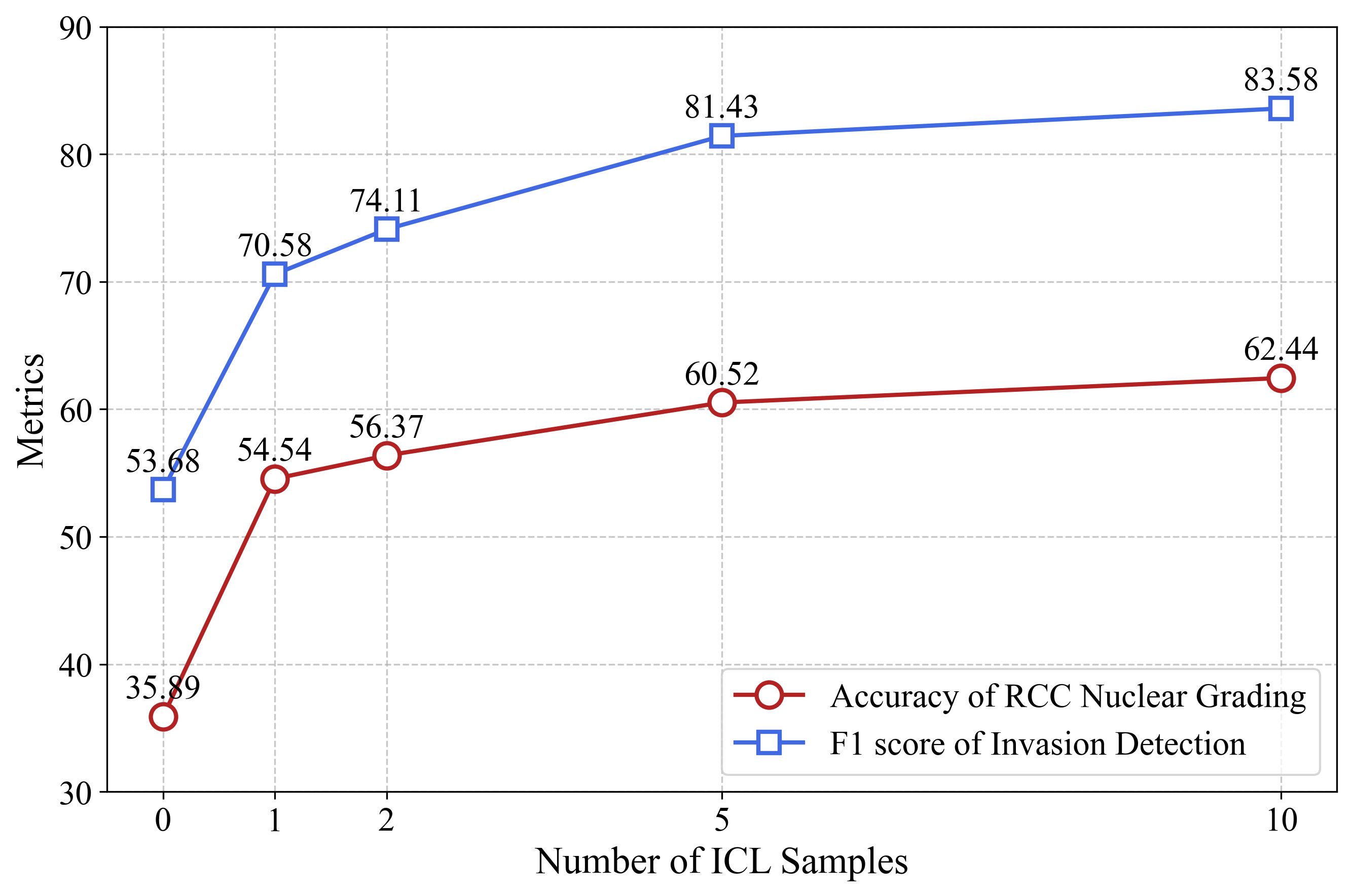}
  \caption {Ablation study on the number of ICL samples.}
  \label{fig:num_icl}
\end{figure}

During the evidence-seeking stage, we incorporate the ICL mechanism when assessing nuclear grades in RCC cases and detecting the presence of invasion. In Fig.~\ref{fig:num_icl}, we study the number of ICL samples. A simple 1-sample ICL can bring much improvement both in the accuracy of RCC nuclear grading and the F1 score (mainly the recall rate) of invasion detection. Incorporating more samples consistently improve the performance, but gets saturated with 5, 10, or more samples.  

\section{Conclusion and Discussion}

In this paper, we present PathFound, an agentic multimodal model that frames pathological diagnosis as an iterative, evidence-seeking process rather than a static single-step task. By integrating three functional modules, a slide highlighter, a vision interpreter, and a diagnostic reasoner, PathFound implements a three-stage protocol that supports progressive hypothesis refinement through repeated visual inspection, evidence collection, and reasoning. Experimental results across multiple foundational models suggest that such an evidence-seeking paradigm can improve diagnostic accuracy and robustness, indicating that many pathological tasks benefit from multi-step reasoning and intermediate evidence aggregation.  PathFound demonstrates competitive performance across most settings, highlighting the role of domain-specific adaptations, including prototype-guided highlighting, staged fine-tuning of vision-language models, and explicit reasoning activation via reinforcement learning. From a broader perspective, two aspects merit further discussion: (1) the relationship between PathFound and existing agentic pathology models, and (2) potential directions for future development. 

Most existing agentic pathology systems primarily focus on an inner observation loop, emphasizing how to systematically examine a given slide for a given task (typically deriving a diagnosis) through plausible and complex navigation. In these settings, the diagnostic objective remains fixed, and the agent iteratively refines its understanding of visual evidence within that scope. By contrast, PathFound introduces an outer diagnostic reasoning loop in which slide observation is driven by evolving diagnostic hypotheses. Rather than repeatedly examining the slide under a static objective, the system dynamically adjusts its visual queries and evidence requirements as the diagnostic process progresses. In addition, PathFound actively requests external information beyond slides, such as additional test results, to further refine diagnostic decisions. These two forms of iteration are complementary: inner observational loops can enhance fine-grained visual understanding, while an outer diagnostic loop can orchestrate hypothesis-driven diagnosis and coordinate heterogeneous sources of evidence to support more consistent and interpretable conclusions. 

Despite the encouraging results, several limitations and open questions remain. First, although PathFound is evaluated across a range of pathological tasks, additional clinical scenarios should be explored to better assess its generality. Closely related to this is the need for more flexible interfaces that can accommodate newly introduced diagnostic settings. At present, while the system supports pan-cancer screening, the fine-grained findings rely on pre-defined toolkits. Extending the framework to previously unseen disease entities with minimal prior information, such as a small number of reference images and brief textual descriptions, remains an important challenge. One possible direction is to leverage lightweight vision-language models that can discover novel visual patterns under weak textual supervision. Second, the current implementation combines a vision-language model with a separate large language model to facilitate explicit reasoning and interpretable outputs. While effective, this design also reflects the current limitations of standalone vision-language models in complex diagnostic reasoning. Further improving the reasoning capabilities of vision-language models themselves, as well as integrating specialized agentic observers into the inner assessment loop, may offer a more unified and efficient solution in future work.

\appendix
\section{Implement Details}
\label{app:imple}
\subsection{Toolkit Building}
Specifically, for the Pan-cancer Abnormality Toolkit, we aggregate tumor and normal regions across 32 distinct cancer types to form the baseline prototypes under both 10× and 20× magnifications. For each type, we utilize 20 slides annotated by experts to develop the template. 

For the specific tasks, we curate fine-grained toolkits that focus on subtle pathological attributes, such as nuclear grade in renal cell carcinoma, Gleason grade in prostate adenocarcinoma, and lymphovascular invasion. Specifically, to construct prototypes for entity-level nuclear grading, we collect 14/83/123/57 images of 20× magnification from TCGA annotations for grades 1 to 4, respectively. Note that, during the re-observation of certain types of RCC, we replace the original randomly selected patches with patches similar to augmented sub-prototypes for each category using $K$-means with $K=4$ to support a more detailed comparison. 
For region-based Gleason grading, we use the annotated regions of Gleason patterns 3-5, Normal, and Stroma from the training set of AGGC 2022 Subset 1 and construct the corresponding prototypes at 20× magnification. Given the heterogeneity of the tissues, we apply $K$-means clustering with $K=2$ to augment the number of prototypes per category. The Gleason patterns 3 to 5 are used in the grading task. 
For invasion detection, we construct prototypes at 5×, 10×, and 20× magnifications for four entity categories, namely vessels with invasion, vessels without invasion, nerves with invasion, and nerves without invasion. For each category, we collect 20 images at the corresponding magnification from TCGA annotations to construct templates. 

\section{Prompts in Use}
\label{app:prompt}

Here is the prompt used in VLM during the Initial Diagnosis stage.

\begin{tcblisting}{
    colback=gray!10,  % 背景色
    colframe=black, % 边框
    title=Vision Interpreter (General fashion),  % 标题
    fonttitle=\bfseries\color{white},     % 标题加粗
    coltitle=black,  % 标题文字颜色
    breakable, % 允许跨页
    listing only,    % 原样显示，不解析 LaTeX 命令
    enhanced,
    sharp corners
}
<Background information>
Given the background information, please analyze the Regions of Interest (RoIs) extracted from the same slide. Perform a comprehensive analysis of the major morphological features of the lesions depicted in these images. Please do not analyze each image individually, but instead focus on an overall evaluation of the tissue. Ignore any watermarks, text, or markings within the images, and focus solely on the tissue and cellular morphology. Please structure your output as follows:

1. Overall Tissue Structure Description:
- Describe the general arrangement, morphology, and distribution of the tissue.
- Identify any structural disorganization, tumor-like clusters, necrotic areas, inflammatory regions, etc.

2. Cellular Morphological Features:
- Describe the size and shape of cells.
- Examine nuclear-to-cytoplasmic ratio, nuclear size, chromatin distribution, nucleoli, mitotic figures, etc.
- Identify any signs of atypia, proliferative activity, etc.
- Identify any special cell types (e.g., lymphocytes, neutrophils, plasma cells, macrophages, foreign-body giant cells, multinucleated cells, etc.) and briefly describe their quantity and distribution.
- Describe any distinct cellular patterns or distributions, such as papillary pattern, glandular pattern and so on.

3. Special Structures or Pathological Changes:
- Mention any specific histological features, such as mitotic figures, inclusion bodies, or extracellular deposits.
- Identify and describe any pathological changes, such as the presence of tumors, calcification, necrosis, inflammation, invasion, or infections.

\end{tcblisting}

Here is the prompt used in VLM during the Evidence Seeking stage with an ICL mechanism.

\begin{tcblisting}{
    colback=gray!10,  % 背景色
    colframe=black, % 边框
    title=Vision Interpreter (ICL-fashion),  % 标题
    fonttitle=\bfseries\color{white},     % 标题加粗
    coltitle=black,  % 标题文字颜色
    breakable, % 允许跨页
    listing only,    % 原样显示，不解析 LaTeX 命令
    enhanced,
    sharp corners
}
Determine if Image A and Image B represent the same tissue or abnormality types. Please answer \"Yes\" or \"No\". \nAnswer the question using a single word or phrase. 

\end{tcblisting}

Here is the prompt used in the exploratory reasoning.

\begin{tcblisting}{
    colback=gray!10,  % 背景色
    colframe=black, % 边框
    title=Exploratory Reasoning prompt,  % 标题
    fonttitle=\bfseries\color{white},     % 标题加粗
    coltitle=black,  % 标题文字颜色
    breakable, % 允许跨页
    listing only,    % 原样显示，不解析 LaTeX 命令
    enhanced,
    sharp corners
}

system:
You are Qwen, created by Alibaba Cloud. You are a helpful assistant. A conversation between User and Assistant. The user asks a question, and the Assistant solves it. The assistant first drafts the reasoning process (inner monologue) until it has derived the final answer with full confidence. It then provides a self-contained summary of the thoughts, i.e., keeping succinct but containing all the critical steps needed to reach the conclusion. It should use Markdown and Latex to format the response. Write both the thoughts and summary in the same language as the task posed by the user.\n\n The thinking process must follow the template below (You should **include and only include one** pair of <think></think> and <answer></answer> tags in your response): \n<think>\n The thoughts or/and draft, like working through an exercise on scratch paper. Be as casual and as long as necessary until it is confident to generate a correct answer.\n</think>\n\n<answer>\n Here, provide a concise summary that reflects the reasoning process and presents a clear final answer to the user.\n</answer>\n

user:
I need you to act as a professional pathologist. After carefully considering the given information, infer the possible differential diagnoses. Then, based on these differential diagnoses, suggest additional information that needs to be provided to rule out certain possibilities. Specifically:  
1. First, you need to carefully analyze the given information, which mainly includes case background information, previous examination items, morphological descriptions of pathological sections, etc. Summarize the evidence points related to the diagnosis from this information. 
2. Based on the given information, analyze what the possible differential diagnoses are and determine whether they are consistent with the given information. Note: These differential diagnoses should be as broad and accurate as possible (broad means considering less common diagnostic possibilities, and accurate means the listed differential diagnoses should not conflict with most of the background information).  
3. According to the listed differential diagnoses, propose the further examination items. You need to specify the exact antigen-antibody, staining type, or molecular type. If the existing information is sufficient to confirm a specific disease, only output that disease and leave the additional examination items blank.  
4. Particularly, when the diagnosis involves renal cell carcinoma, you can call the relevant tools to further observe the original images and collect evidence. The tools include: renal clear cell observation tool (tool-ccRCC), renal chromophobe cell observation tool (tool-chRCC), renal papillary cell observation tool (tool-pRCC), and Furhman nuclear grade observation tool (tool-Nuclear). When the diagnosis involves prostate adenocarcinoma, you can call the tool (tool-Gleason) to evaluate the Gleason score. When necessary, you can call the invasion detection tool (tool-invasion) to detect any lymphovascular or perineural invasion. 
5. Finally, summarize the possible differential diagnoses and the required additional examination items in a given format. When summarizing the differential diagnoses, you need to rank the more likely diagnoses higher.
6. When the given information is sufficient to confirm a specific disease, you can directly output the diagnosis result and skip the differential diagnosis analysis. In the format of the differential diagnosis analysis, only keep the final diagnosis and leave other items blank.
7. When outputting, you need to structure your response and strictly follow the format requirements: 
Differential Diagnoses: \DiffList{Diagnosis 1, Diagnosis 2, ...}
Further Examination Items: \ExamList{Item 1, Item 2, ...}
Further Observation Tool Calls: \ToolCallList{Tool 1, Tool 2, ...}

The following is the case information:
<Case Information>
\end{tcblisting}

Here is the prompt used in the definitive reasoning.

\begin{tcblisting}{
    colback=gray!10,  % 背景色
    colframe=black, % 边框
    title=Definitive Reasoning prompt,  % 标题
    fonttitle=\bfseries\color{white},     % 标题加粗
    coltitle=black,  % 标题文字颜色
    breakable, % 允许跨页
    listing only,    % 原样显示，不解析 LaTeX 命令
    enhanced,
    sharp corners
}

user:
Now the results of the further examinations have come out. I need you to:
1. First, check the "Case Information" and the first-round diagnosis to sort out the previous diagnostic chain of thought and related conclusions.
2. Then, check the "Results of Further Examinations" and the "Results of Further Observations". The additional test results may not fully match the items requested in the initial diagnosis. Based on the available test results, you need to conduct further differential analysis, and give the final diagnosis. The results of further observations are the results obtained after the first-round tool call. Note: You are completely entitled to overturn the initial diagnostic approach and provide a diagnosis based on the current information after obtaining more data.
3. The final diagnosis must be output in the specified format, i.e., \boxed{Diagnosis Name}
4. If you are still unable to determine a specific disease at this time, continue to output possible diseases (as concise as possible). Format: \DiffList{Diagnosis1, Diagnosis2}

Here is the information:
Results of Further Examinations: <Exam Results>
Results of Further Observations: <Further Observations>
\end{tcblisting}

\section{Case Study}
\label{app:case}

We present three cases in this section. The first case is a clear cell RCC from TCGA to demonstrate the diagnostic process, which involves further observation of cell types and nuclear grades. The second is a stomach adenocarcinoma from TCGA, confirmed to have lymphatic invasion, which is used to demonstrate PathFound's invasion detection capability. The third is a complex case from Pathology Outlines\footnote{https://www.pathologyoutlines.com}, which contains a longer diagnostic path to refine the final diagnosis. 

\subsection{TCGA-B0-4824 (Clear Cell RCC)}

Here we present a complete diagnosis by PathFound. The case concerns renal clear cell carcinoma with nuclear grade 3.

\begin{figure}[ht]
  \centering
  \includegraphics[width=\linewidth]{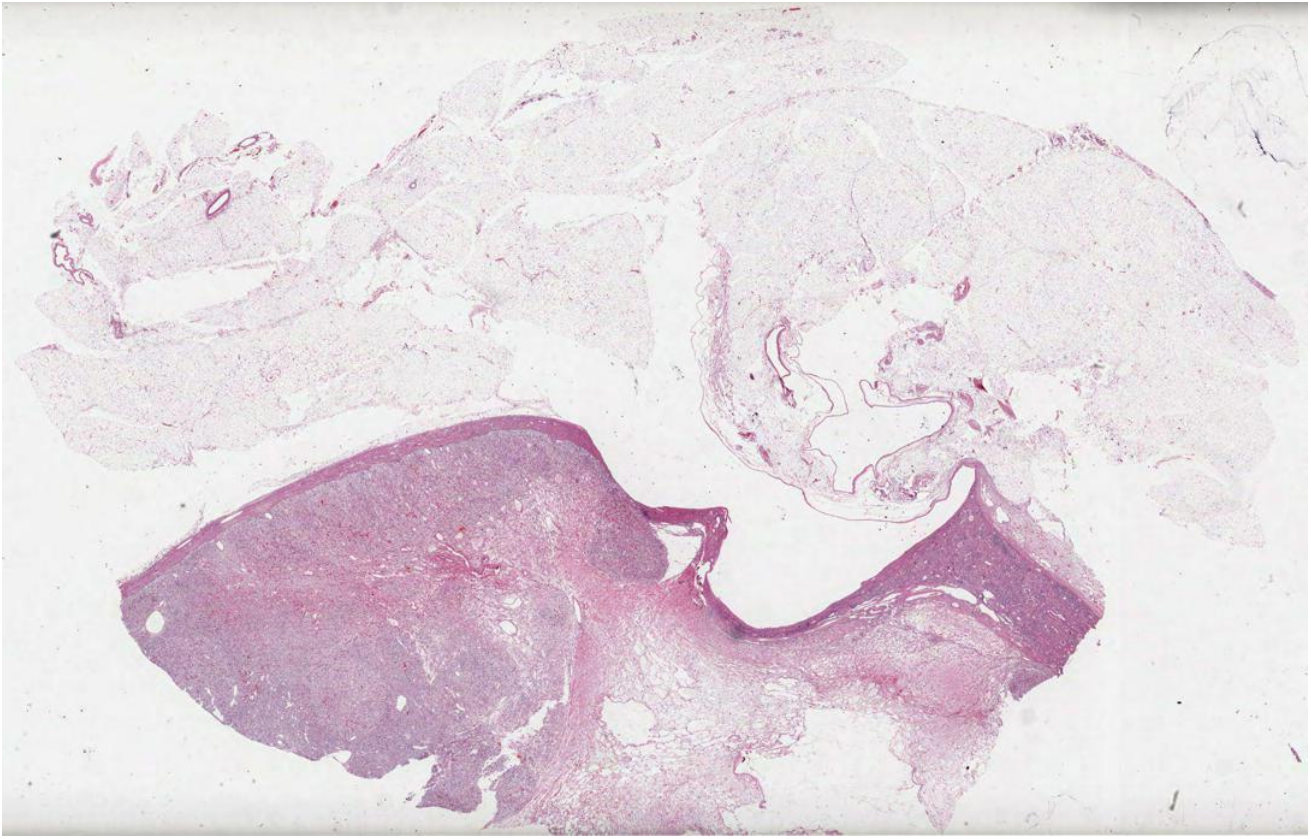}
  \caption {Original slide of TCGA-B0-4824 . }
  \label{fig:ori_wsi}
\end{figure}

We first present the preliminary information, including the basic information and the original slide (Fig.~\ref{fig:ori_wsi}).

\begin{tcolorbox}[
    colback=gray!10,      
    colframe=black,       
    title=Basic Information,
    fonttitle=\bfseries\color{white},  
    coltitle=black,       
    boxrule=0.5pt,        
    arc=2mm,              
    top=2mm, bottom=2mm, left=2mm, right=2mm,
    breakable,
]

\textbf{Case Information:} The slide is from the left kidney, with a tumor of 3.5 cm. 
\end{tcolorbox}

\begin{figure}[ht]
  \centering
  \includegraphics[width=\linewidth]{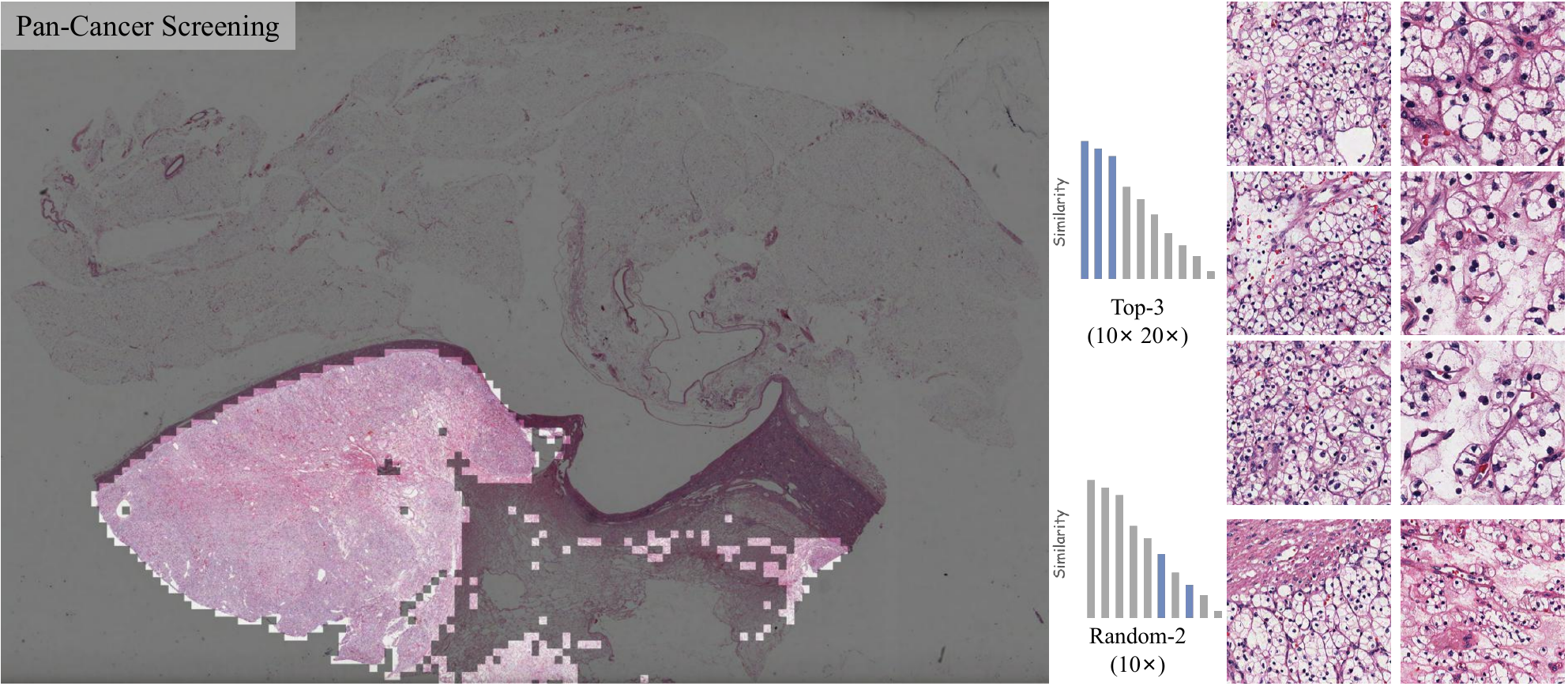}
  \caption {Highlighted regions and sample RoIs during the initial diagnosis stage. }
  \label{fig:wsi_1}
\end{figure}

We then present the highlighted regions and sampled RoIs (Fig.~\ref{fig:wsi_1}) from the initial diagnosis, along with the general microscopic findings.

\begin{tcolorbox}[
    colback=gray!10,      
    colframe=black,       
    title=Microscopic Findings (Initial Diagnosis),
    fonttitle=\bfseries\color{white},  
    coltitle=black,       
    boxrule=0.5pt,        
    arc=2mm,              
    top=2mm, bottom=2mm, left=2mm, right=2mm,
    breakable,
]

Microscopic observation:

\textbf{Overall Tissue Structure Description:}

The tissue exhibits a disorganized structure characterized by nests and sheets of cells separated by thin vascular septa. Small acinar or tubular-like structures are also present, replacing the normal renal parenchyma. Cells display polygonal shapes with clear or pale eosinophilic cytoplasm and distinct cell borders.

\vspace{0.5em}

\textbf{Cellular Morphological Features:}

Cells are polygonal with a high nucleus-to-cytoplasm ratio. The nuclei are round to oval, centrally located, and contain finely granular chromatin. Prominent nucleoli are observed. There is no significant mitotic activity, and the cells exhibit some pleomorphism but maintain a relatively consistent appearance.

\vspace{0.5em}

\textbf{Special Structures or Pathological Changes:}

No mitotic figures, necrosis, or significant inflammatory infiltrate are present. The tumor appears well-demarcated without clear evidence of invasion into surrounding tissues.
\end{tcolorbox}

PathFound's diagnostic reasoner provides the first reasoning based on previous information as follows:

\begin{tcolorbox}[
    colback=gray!10,      
    colframe=black,       
    title=Diagnostic Reasoning (Initial Diagnosis),
    fonttitle=\bfseries\color{white},  
    coltitle=black,       
    boxrule=0.5pt,        
    arc=2mm,              
    top=2mm, bottom=2mm, left=2mm, right=2mm,
    breakable,
]
\textless think\textgreater

First, according to the case information, we can determine that this is a tumor arising from the left kidney, with a diameter of 3.5\,cm. Microscopic examination of the tissue architecture shows that the tissue displays an irregular structure composed of nests and sheets of cells, which are separated by thin vascular septa. Some small acinar or tubular structures are also present, replacing the normal renal parenchyma. The cells are polygonal, with clear or pale eosinophilic cytoplasm and distinct cell borders.

The cytomorphological features show that the cells are polygonal with a high nuclear-to-cytoplasmic ratio. The nuclei are round or oval, centrally located, and contain finely granular chromatin with visible nucleoli. There is no significant mitotic activity. The cells exhibit some degree of pleomorphism, but the overall appearance is relatively uniform.

Regarding special structures or pathological changes, no mitotic figures, necrosis, or significant inflammatory infiltrates are identified. The tumor is well circumscribed, with no obvious evidence of invasion into the surrounding tissues.

Based on this information, the preliminary differential diagnosis may include:
\begin{enumerate}
  \item Clear cell renal cell carcinoma (ccRCC)
  \item Chromophobe renal cell carcinoma (chRCC)
  \item Papillary renal cell carcinoma (pRCC)
\end{enumerate}

Next, we need to further exclude or confirm these diagnoses. In order to distinguish among these possible diagnoses, additional examinations are required, such as immunohistochemical staining. For example, PAX8 and CD10 are very useful for differentiating among these types of renal cell carcinoma.

When summarizing the differential diagnosis, diagnoses with a higher likelihood should be listed first. Based on the available information, ccRCC is more likely, because the described cellular pleomorphism and high nuclear-to-cytoplasmic ratio are associated with ccRCC. chRCC and pRCC are also possible, but further studies are needed to distinguish them.

Therefore, the preliminary differential diagnosis list may be: ccRCC, chRCC, and pRCC.

The additional examination items may include:
\begin{itemize}
  \item Immunohistochemical staining: PAX8, CD10, CK7, CK20
\end{itemize}

The review tool calls may include:
\begin{itemize}
  \item tool-ccRCC
  \item tool-chRCC
  \item tool-pRCC
  \item tool-Nuclear
\end{itemize}

Next, I will organize this information and provide the final answer.

\textless /think\textgreater

\textless answer\textgreater

\textbf{Differential diagnosis:}  

\textbackslash DiffList\{\textcolor{green!60!black}{Clear cell renal cell carcinoma (ccRCC)}, Chromophobe renal cell carcinoma (chRCC), Papillary renal cell carcinoma (pRCC)\}

\vspace{0.5em}

\textbf{Further Examination Items:}  

\textbackslash ExamList\{Immunohistochemical staining PAX8, Immunohistochemical staining CD10, Immunohistochemical staining CK7, Immunohistochemical staining CK20\}

\vspace{0.5em}

\textbf{Further Observation Tool Calls:}  

\textbackslash ToolCallList\{tool-ccRCC, tool-chRCC, tool-pRCC, tool-Nuclear\}

\textless /answer\textgreater

\end{tcolorbox}

\begin{figure}[ht]
  \centering
  \includegraphics[width=\linewidth]{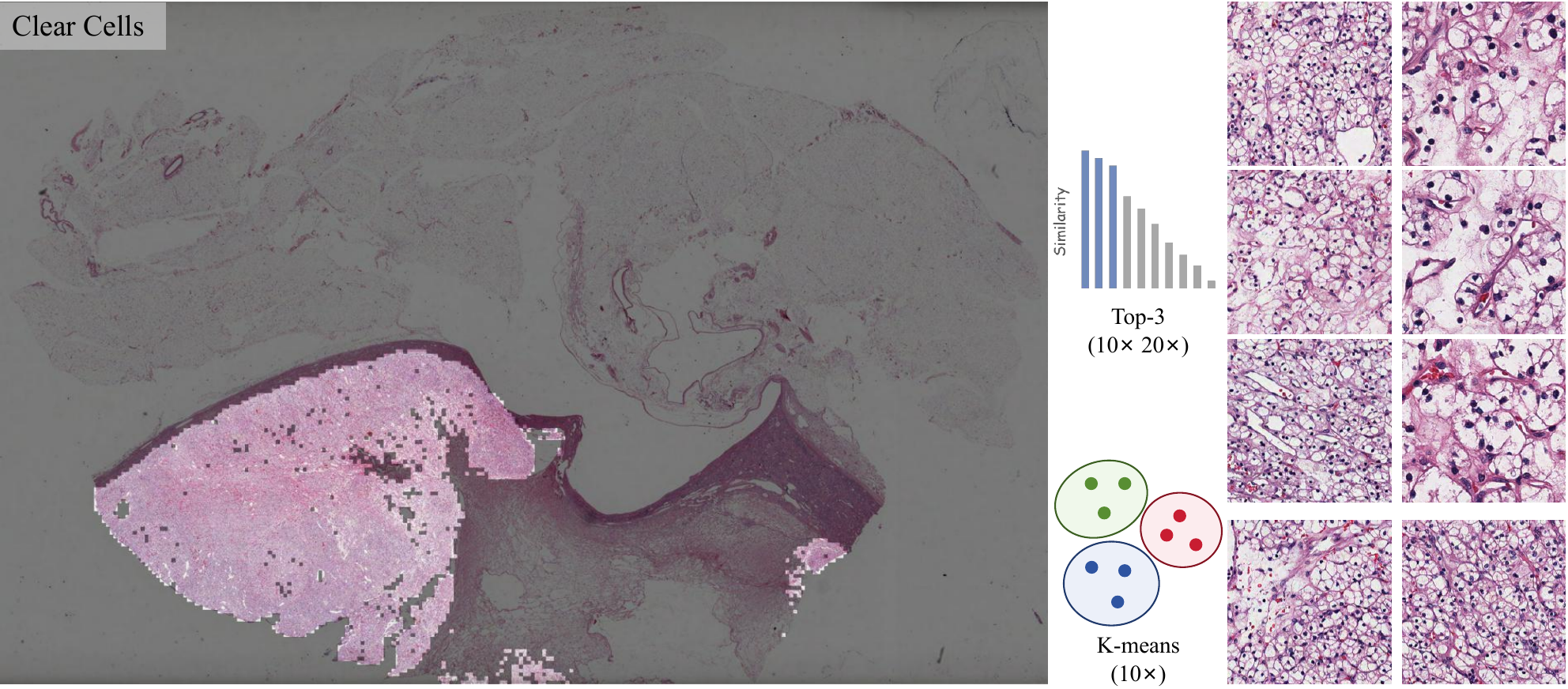}
  \caption {Highlighted regions and sampled RoIs when calling the specific observation tool of clear cells. }
  \label{fig:roi_cc}
\end{figure}

\begin{figure}[ht]
  \centering
  \includegraphics[width=\linewidth]{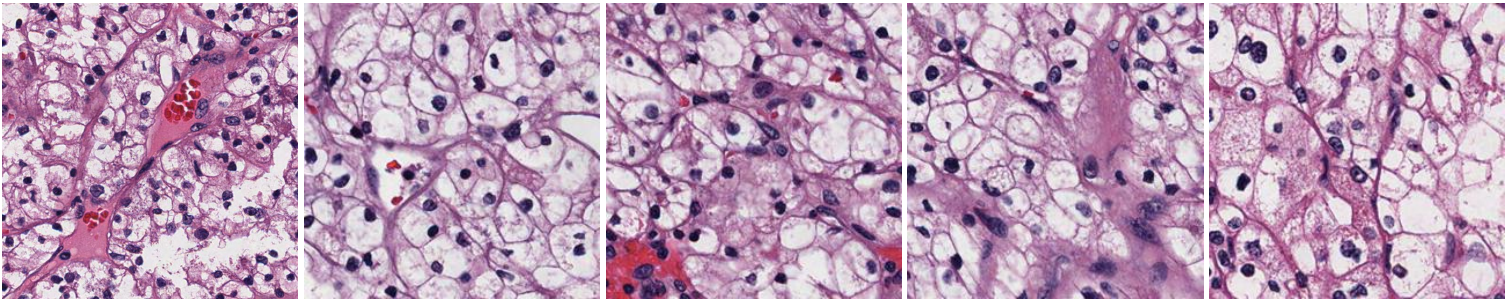}
  \caption {Sampled RoIs when identifying cells with nuclear grade 3. }
  \label{fig:roi_nuclear}
\end{figure}

Guided by the plan of further acquisition, the slide highlighter and the interpreter are re-triggered to provide a closer look at the original slide. We present the highlighted regions and sampled RoIs to identify clear cells (Fig.~\ref{fig:roi_cc}) and cells with nuclear grade 3 (Fig.~\ref{fig:roi_nuclear}). Also, we simulate the results of request tests. The additional evidence is listed as follows.

\begin{tcolorbox}[
    colback=gray!10,      
    colframe=black,       
    title=Additional Evidence,
    fonttitle=\bfseries\color{white},  
    coltitle=black,       
    boxrule=0.5pt,        
    arc=2mm,              
    top=2mm, bottom=2mm, left=2mm, right=2mm,
    breakable,
]

\textbf{Additional Observations:} 

ccRCC: positive \qquad chRCC: negative \qquad pRCC: negative \qquad Nuclear grade: 3

\textbf{Additional Test Results:} 

PAX8: Positive \qquad CD10: Postive \qquad CK7: Negative \qquad CD20: Negative 

\end{tcolorbox}

Then we initiate the final decision.

\begin{tcolorbox}[
    colback=gray!10,      
    colframe=black,       
    title=Diagnostic Reasoning (Final Decision),
    fonttitle=\bfseries\color{white},  
    coltitle=black,       
    boxrule=0.5pt,        
    arc=2mm,              
    top=2mm, bottom=2mm, left=2mm, right=2mm,
    breakable,
]
\textless think\textgreater

First, the previous case information and first-round diagnosis are reviewed:

\begin{itemize}
  \item The case information shows a tumor located in the left kidney, measuring 3.5\,cm in diameter. Microscopic examination reveals polygonal cells with clear or pale eosinophilic cytoplasm, a high nuclear-to-cytoplasmic ratio, round or oval centrally located nuclei containing finely granular chromatin and prominent nucleoli. There is no significant mitotic activity. The cells show some pleomorphism, but the overall appearance is relatively uniform. No mitotic figures, necrosis, or significant inflammatory infiltrates are identified. The tumor is well circumscribed, with no obvious invasion into surrounding tissues.
  
  \item The preliminary differential diagnoses include: clear cell renal cell carcinoma (ccRCC), chromophobe renal cell carcinoma (chRCC), and papillary renal cell carcinoma (pRCC).
  
  \item The additional examination results are as follows:
  \begin{itemize}
    \item PAX8: positive
    \item CD10: positive
    \item CK7: negative
    \item CK20: negative
  \end{itemize}
  
  \item The additional observations are as follows:
  \begin{itemize}
    \item Renal clear cells: positive
    \item Renal chromophobe cells: negative
    \item Renal papillary cells: negative
    \item Nuclear grade: 3
  \end{itemize}
\end{itemize}

Based on the additional examination and observation results:
\begin{itemize}
  \item PAX8 positivity is a general marker for renal cell carcinoma.
  \item CD10 positivity supports clear cell renal cell carcinoma (ccRCC).
  \item CK7 and CK20 negativity exclude some other possibilities.
  \item Renal clear cells positive further support ccRCC.
  \item Renal chromophobe and papillary cells negative exclude chRCC and pRCC.
  \item Nuclear grade 3 indicates a certain degree of malignancy.
\end{itemize}

Integrating the above information, the most likely diagnosis is clear cell renal cell carcinoma (ccRCC).

Therefore, the final diagnosis should be clear cell renal cell carcinoma (ccRCC), nuclear grade 3.

\textless /think\textgreater

\textless answer\textgreater

\textbackslash boxed\{\textcolor{green!60!black}{Clear cell renal cell carcinoma (ccRCC), nuclear grade 3}\}

\textless /answer\textgreater

\end{tcolorbox}

\subsection{TCGA-VQ-A91K (Stomach Adenocarcinoma with Lymphatic Invasion)}
Here we present another complete diagnosis by PathFound. The case concerns stomach adenocarcinoma with lymphatic invasion.

\begin{figure}[ht]
  \centering
  \includegraphics[width=\linewidth]{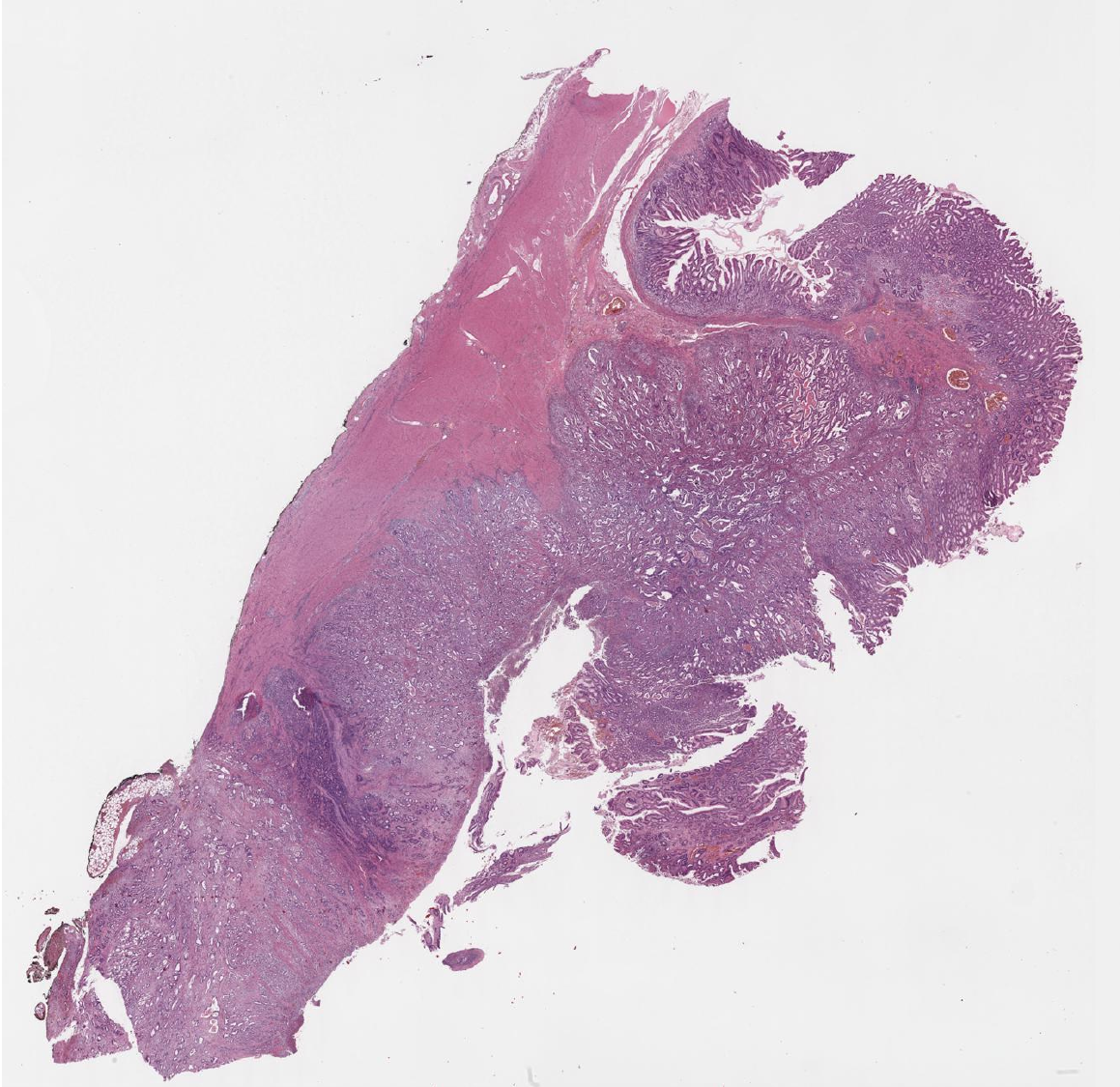}
  \caption {Original slide of TCGA-VQ-A91K. }
  \label{fig:ori_wsi_2}
\end{figure}

We first present the preliminary information, including the basic information and the original slide (Fig.~\ref{fig:ori_wsi_2}).

\begin{tcolorbox}[
    colback=gray!10,      
    colframe=black,       
    title=Basic Information,
    fonttitle=\bfseries\color{white},  
    coltitle=black,       
    boxrule=0.5pt,        
    arc=2mm,              
    top=2mm, bottom=2mm, left=2mm, right=2mm,
    breakable,
]

This pathology slide describes the findings from a stomach (antrum) tumor resection. The tumor with a size of 2.8 cm located in the pyloric region and duodenum.

\end{tcolorbox}

\begin{figure}[ht]
  \centering
  \includegraphics[width=\linewidth]{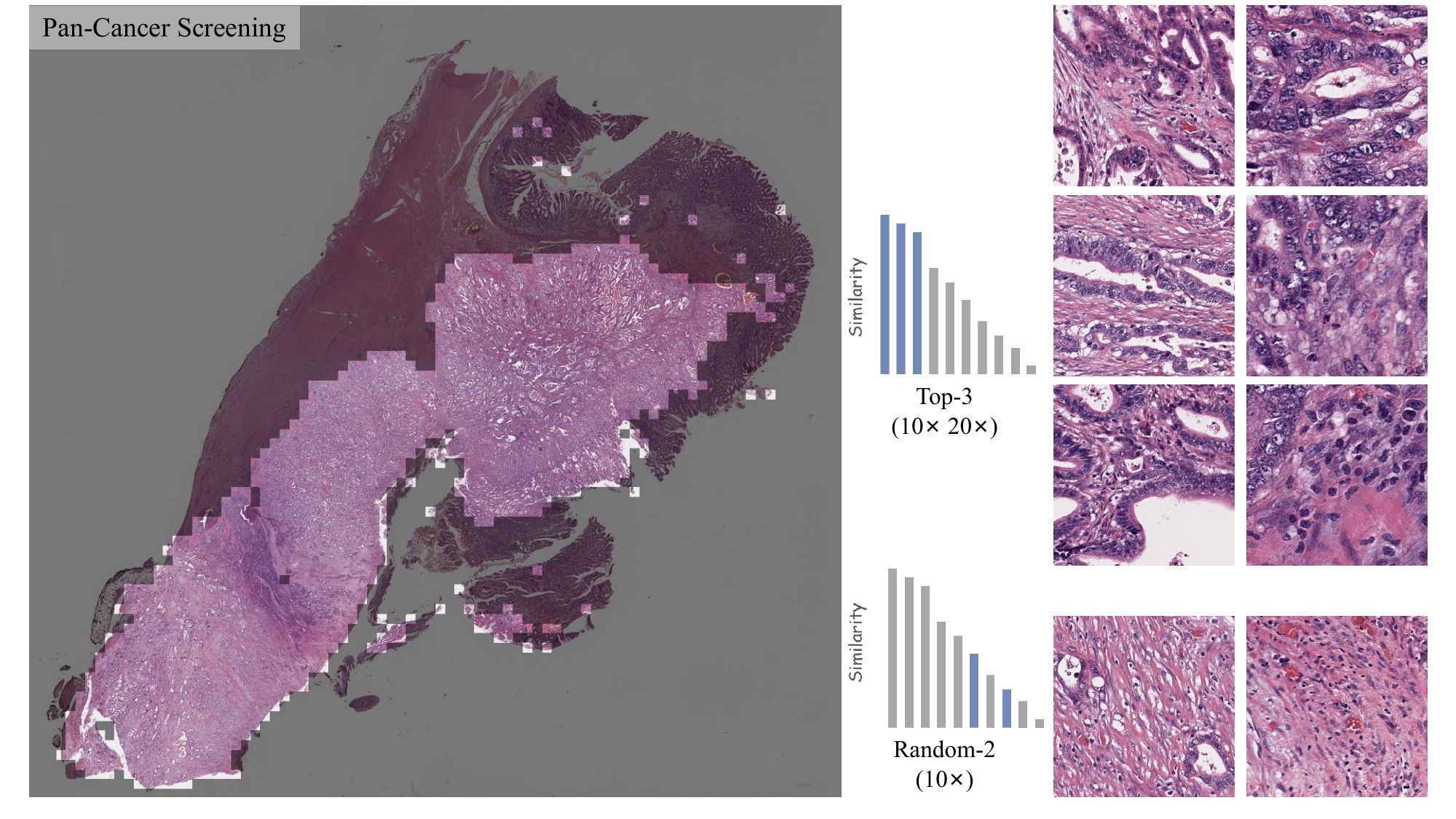}
  \caption {Highlighted regions and sampled RoIs during the initial diagnosis stage of TCGA-VQ-A91K.}
  \label{fig:pancancer_2}
\end{figure}

We then present the highlighted regions and sampled RoIs (Fig.~\ref{fig:pancancer_2}) from the initial diagnosis, along with the general microscopic findings.

\begin{tcolorbox}[
    colback=gray!10,      
    colframe=black,       
    title=Microscopic Findings (Initial Diagnosis),
    fonttitle=\bfseries\color{white},  
    coltitle=black,       
    boxrule=0.5pt,        
    arc=2mm,              
    top=2mm, bottom=2mm, left=2mm, right=2mm,
    breakable,
]
Microscopic observation:

\textbf{Section 1: Overall Tissue Structure Description}

The stomach wall shows extensive replacement by malignant epithelium, disrupting the normal architecture. The tumor consists of infiltrative tumor buds with irregular, back-to-back, and complex outpouching glands. These glands exhibit various patterns, ranging from simple tubular to cribriform and papillary, indicating a degree of anaplasia.

The stroma surrounding the tumor is markedly desmoplastic—dense and fibrous, with a reactive fibroblastic response. Scattered throughout the inflammatory infiltrate are numerous lymphocytes, plasma cells, and occasional eosinophils. No significant necrosis is observed, but there is a high cellular density and rapid replication rate indicated by the high mitotic index.

\vspace{0.5em}

\textbf{Section 2: Cellular Morphological Features}

The tumor cells show marked cytologic atypia. Nuclei are significantly enlarged, often two to five times the size of benign gastric epithelial nuclei. The nucleus-to-cytoplasm ratio is increased, and the chromatin is coarse and hyperchromatic. Nuclear membranes are irregular, and chromatin clumping gives the nucleus a mottled appearance. Binucleation and multinucleation are common. Prominent nucleoli are frequently present.

\vspace{0.5em}

\textbf{Section 3: Special Structures or Pathological Changes}

- \textbf{Architectural Complexity:} The tumor displays complex glandular configurations, including cribriform structures and irregular papillary projections.

- \textbf{Proliferative Activity:} Numerous mitotic figures are easily identified.

- \textbf{Desmoplastic Stroma:} Dense, fibrous stroma surrounds the glandular structures.

- \textbf{Inflammatory Infiltrate:} A mixed population of inflammatory cells, including lymphocytes and plasma cells, is present within the tumor and surrounding stroma.
\end{tcolorbox}

PathFound's diagnostic reasoner provides the first reasoning based on previous information as follows:

\begin{tcolorbox}[
    colback=gray!10,      
    colframe=black,       
    title=Diagnostic Reasoning (Initial Diagnosis),
    fonttitle=\bfseries\color{white},  
    coltitle=black,       
    boxrule=0.5pt,        
    arc=2mm,              
    top=2mm, bottom=2mm, left=2mm, right=2mm,
    breakable,
]

\textless think\textgreater

First, according to the case information, the tumor is located in the antrum of the stomach and the duodenum, measuring 2.8\,cm in size. Microscopic examination reveals that the tumor disrupts the normal gastric wall architecture, showing infiltrative tumor buds with irregular, back-to-back, and complex cystic glands. These glands exhibit a spectrum of architectural patterns ranging from simple tubular to cribriform and papillary forms, indicating a certain degree of atypia. The surrounding stroma is markedly desmoplastic, accompanied by reactive fibroblast proliferation. The inflammatory infiltrate contains abundant lymphocytes, plasma cells, and occasional eosinophils. No obvious necrosis is identified; however, the cellular density is high, with an increased mitotic index.

Cytomorphological features demonstrate that the tumor cells show marked cytological atypia, with significantly enlarged nuclei and an increased nuclear-to-cytoplasmic ratio. The chromatin is coarse and hyperchromatic, the nuclear membranes are irregular, and nucleoli are prominent. These features suggest the possibility of gastric adenocarcinoma. Based on the described complex glandular architecture, proliferative activity, and desmoplastic stroma, the possible differential diagnoses include:
\begin{enumerate}
  \item \textbf{Gastric adenocarcinoma}
  \item \textbf{Gastric neuroendocrine tumor}
  \item \textbf{Gastric lymphoma}
  \item \textbf{Gastric stromal tumor}
\end{enumerate}

Next, it is necessary to further exclude certain possibilities. Gastric lymphoma typically does not show such prominent glandular structures or proliferative activity, whereas gastric stromal tumors usually consist of spindle cells or epithelioid cells with a more abundant stromal component. Gastric neuroendocrine tumors, although they may show atypia, usually do not demonstrate such complex glandular architecture.

Therefore, the most likely differential diagnosis is gastric adenocarcinoma. In order to further exclude other possibilities, the following additional examinations may be required:
\begin{enumerate}
  \item \textbf{Immunohistochemical studies}: such as CK7, CK20, CDX2, Synaptophysin, and CgA, to distinguish adenocarcinoma from other types of tumors.
  \item \textbf{Ki-67 staining}: to assess the proliferative index.
  \item \textbf{HER2 testing}: to evaluate HER2 status, which is of important significance for the treatment of gastric adenocarcinoma.
\end{enumerate}

As the case information is mainly focused on a gastric tumor, there is no need to invoke review tools related to renal cell carcinoma or prostate carcinoma. It is necessary to invoke the review tool of invasion.

The summary is as follows.

\textbf{Differential diagnosis:} gastric adenocarcinoma, gastric neuroendocrine tumor, gastric lymphoma, and gastric stromal tumor.

\textbf{Additional examination items:} immunohistochemistry (CK7, CK20, CDX2, Synaptophysin, CgA), Ki-67 staining, and HER2 testing.

\textbf{Tool call review:} tool-invasion.

\textless /think\textgreater

\textless answer\textgreater

\textbf{Differential diagnosis:}  

\textbackslash DiffList\{\textcolor{green!60!black}{Gastric adenocarcinoma}, Gastric neuroendocrine tumor, Gastric lymphoma, Gastric stromal tumor\}

\vspace{0.5em}

\textbf{Further Examination Items:}  

\textbackslash ExamList\{Immunohistochemistry (CK7, CK20, CDX2, Synaptophysin, CgA), Ki-67 staining, HER2 testing\}

\vspace{0.5em}

\textbf{Further Observation Tool Calls:}  

\textbackslash ToolCallList\{tool-invasion\}

\textless /answer\textgreater

\end{tcolorbox}

\begin{figure}[ht]
  \centering
  \includegraphics[width=\linewidth]{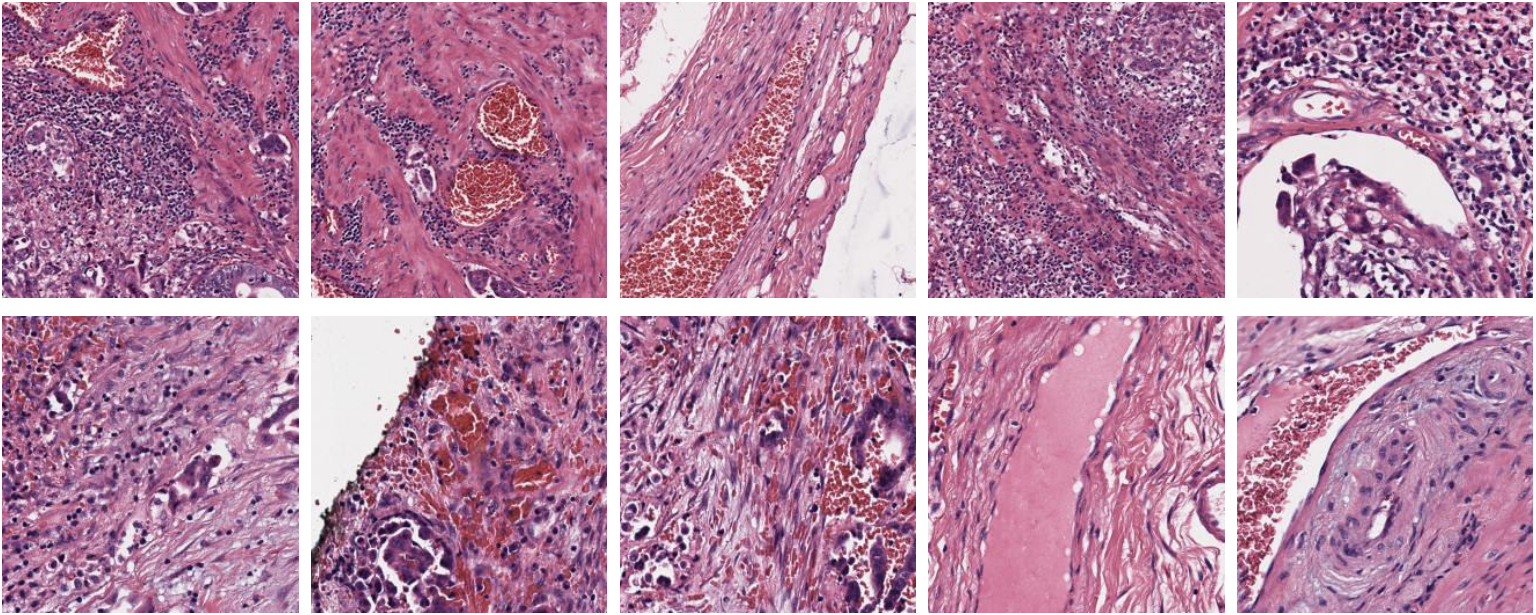}
  \caption {Sampled RoIs when calling the specific observation tool of invasion. }
  \label{fig:invasion}
\end{figure}

Guided by the plan of further acquisition, the slide highlighter and the interpreter are re-triggered to provide a closer look at the original slide. We present sampled RoIs to identify the invasion (Fig.~\ref{fig:invasion}). Also, we simulate the results of request tests. The additional evidence is listed as follows. 

\begin{tcolorbox}[
    colback=gray!10,      
    colframe=black,       
    title=Additional Evidence,
    fonttitle=\bfseries\color{white},  
    coltitle=black,       
    boxrule=0.5pt,        
    arc=2mm,              
    top=2mm, bottom=2mm, left=2mm, right=2mm,
    breakable,
]
\textbf{Additional observations:}

Invasion: detected

\textbf{Additional examination results:}
\begin{itemize}
  \item CK7: positive
  \item CK20: positive
  \item CDX2: positive
  \item Synaptophysin: negative
  \item CgA: negative
  \item Ki-67 staining: increased proliferative index
  \item HER2 testing: negative
\end{itemize}

\end{tcolorbox}

\begin{tcolorbox}[
    colback=gray!10,      
    colframe=black,       
    title=Diagnostic Reasoning (Final Decision),
    fonttitle=\bfseries\color{white},  
    coltitle=black,       
    boxrule=0.5pt,        
    arc=2mm,              
    top=2mm, bottom=2mm, left=2mm, right=2mm,
    breakable,
]
\textless think\textgreater
First, the previous diagnostic reasoning and conclusions are reviewed. According to the initial case information, the preliminary diagnoses included gastric adenocarcinoma, gastric neuroendocrine tumor, gastric lymphoma, and gastric stromal tumor. The additional examination results included immunohistochemical studies and Ki-67 staining.

\begin{enumerate}
  \item \textbf{Immunohistochemical results:}
  \begin{itemize}
    \item CK7: positive
    \item CK20: positive
    \item CDX2: positive
    \item Synaptophysin: negative
    \item CgA: negative
  \end{itemize}

  \item \textbf{Ki-67 staining}: increased proliferative index
  \item \textbf{HER2 testing}: negative
  \item \textbf{Invasion}: detected
\end{enumerate}

These results further help to exclude certain diagnoses:
\begin{itemize}
  \item Both \textbf{Synaptophysin} and \textbf{CgA} are negative, which excludes the possibility of gastric neuroendocrine tumor.
  \item Positivity for \textbf{CK7} and \textbf{CK20}, together with \textbf{CDX2} positivity, supports the diagnosis of gastric adenocarcinoma.
  \item \textbf{Ki-67} positivity indicates that the tumor has proliferative activity, which is consistent with the features of gastric adenocarcinoma.
  \item \textbf{HER2} negativity excludes HER2-positive subtypes of gastric adenocarcinoma.
\end{itemize}

Therefore, based on the integrated information, the most likely diagnosis is gastric adenocarcinoma \textcolor{green!60!black}{with certain invasion}. Other possibilities, such as gastric lymphoma and gastric stromal tumor, can be excluded under the current information.

\textbf{Final diagnosis:}

\textbackslash boxed\{Gastric adenocarcinoma\}

\textless /think\textgreater

\textless answer\textgreater

\textbackslash boxed\{\textcolor{green!60!black}{Gastric adenocarcinoma}\}

\textless /answer\textgreater

\end{tcolorbox}

\subsection{Pathology Outlines Case 479 (Thymic Carcinoma)} 

To protect the copyright of the images on the website, we only provide the corresponding URL\footnote{https://www.pathologyoutlines.com/caseofweek/case479.htm}, and readers can visit the website to view them at their own discretion. The following demonstrates the whole diagnostic process.

\begin{tcolorbox}[
    colback=gray!10,      
    colframe=black,       
    title=Basic Information,
    fonttitle=\bfseries\color{white},  
    coltitle=black,       
    boxrule=0.5pt,        
    arc=2mm,              
    top=2mm, bottom=2mm, left=2mm, right=2mm,
    breakable,
]

\textbf{Case Information:} A 47 year old Asian woman presented with malaise, weight loss, chest pain and shortness of breath. CT scan showed an anterior mediastinal mass with irregular, ill-defined and infiltrative borders, measuring up to 8.3 x 7.1 x 8.2 cm. A biopsy was performed. 
\end{tcolorbox}

\begin{tcolorbox}[
    colback=gray!10,      
    colframe=black,       
    title=Microscopic Findings (Initial Diagnosis),
    fonttitle=\bfseries\color{white},  
    coltitle=black,       
    boxrule=0.5pt,        
    arc=2mm,              
    top=2mm, bottom=2mm, left=2mm, right=2mm,
    breakable,
]
Microscopic observation:

\textbf{1. Overall Tissue Structure Description:}

- The tumor appears highly cellular and densely packed.

- The cells are arranged in clusters and sheets, with some areas showing a somewhat cohesive pattern, potentially forming glandular or papillary structures, although this is not definitively clear at high magnification.

- There is no significant evidence of organized fibrous stroma, collagen deposition, or dense inflammatory infiltrate visible in the sections shown.

- The tissue lacks well-formed vascular channels or organized muscle layers, indicating a lack of normal organ-specific architecture preservation.

\quad

\textbf{2. Cellular Morphological Features:}

- \textbf{Cell Size and Shape:} The cells are predominantly small to medium-sized with relatively uniform, round to slightly oval nuclei. Some variation exists, but marked pleomorphism is absent.

- \textbf{Nuclear-to-Cytoplasmic Ratio:} The nuclear-to-cytoplasmic ratio is increased; nuclei occupy most of the cell volume, leaving only minimal eosinophilic cytoplasm visible.

- \textbf{Nuclear Size, Chromatin Distribution, Nucleoli:} The nuclei are hyperchromatic, meaning they stain darkly due to increased DNA content. Chromatin appears somewhat finely dispersed rather than uniformly condensed. Small, inconspicuous nucleoli are present within many nuclei.

- \textbf{Proliferative Activity:} No overt mitotic figures are clearly identifiable in these low-power fields.

- \textbf{Atypia:} There is no significant nuclear atypia, such as irregular nuclear contours, vesicular chromatin, or prominent nucleoli.

- \textbf{Special Cell Types:} Lymphocytes, neutrophils, plasma cells, or other immune cells are not readily apparent in the provided images. Some scattered cells appear smaller, possibly representing lymphocytes or tumor cell apoptosis. No macrophages or multinucleated giant cells are evident.

- \textbf{Distinct Cellular Patterns:} The cells are arranged in solid sheets, with some areas showing a loosely cohesive or papillary-like growth pattern, characterized by cells arranged around a central space.

\quad

\textbf{3. Special Structures or Pathological Changes:}

- \textbf{Mitotic Figures:} No overt mitotic figures are present in the examined fields.

- \textbf{Inclusion Bodies/Extracellular Deposits:} No obvious viral inclusions, fungal hyphae, or crystal deposits are identified.

- \textbf{Pathological Changes:} The most significant finding is the marked cellular proliferation forming irregular clusters and sheets. The architectural distortion and cellular density are indicative of a neoplasm.
\end{tcolorbox}

\begin{tcolorbox}[
    colback=gray!10,      
    colframe=black,       
    title=Diagnostic Reasoning (Initial Diagnosis),
    fonttitle=\bfseries\color{white},  
    coltitle=black,       
    boxrule=0.5pt,        
    arc=2mm,              
    top=2mm, bottom=2mm, left=2mm, right=2mm,
    breakable,
]

\textless think\textgreater

First, according to the case information, the patient is a 47-year-old Asian female whose chief complaints are fatigue, weight loss, chest pain, and dyspnea. CT imaging demonstrates an irregular, infiltrative mass located in the anterior mediastinum, measuring approximately $8.3 \times 7.1 \times 8.2\,\mathrm{cm}$.

The biopsy results show that the tissue is highly cellular, with tumor cells arranged in clusters and sheets. In some areas, glandular or papillary structures may be formed, but these are not obvious. The cells are relatively uniform in size and shape, with an increased nuclear-to-cytoplasmic ratio. The nuclear chromatin is finely distributed, and the nucleoli are small and inconspicuous. There is no obvious nuclear atypia or mitotic figures. The cells are arranged in solid sheets, and some areas show a loose or papillary growth pattern.

The preliminary differential diagnosis may include:
\begin{enumerate}
  \item \textbf{Thymic carcinoma / thymoma}: an anterior mediastinal mass with high cellularity, which may form glandular or papillary structures.
  \item \textbf{Lymphoma}: an anterior mediastinal mass with high cellularity, which may show a sheet-like or solid growth pattern.
  \item \textbf{Germ cell tumor}: an anterior mediastinal mass with high cellularity, which may show a sheet-like or solid growth pattern.
  \item \textbf{Neuroendocrine tumor}: an anterior mediastinal mass with high cellularity, which may show a sheet-like or solid growth pattern.
\end{enumerate}

Next, it is necessary to further exclude some diagnoses with lower probability, such as sarcoma or other types of tumors, because the case information does not mention obvious fibrous or muscular structures. At the same time, it is necessary to consider whether other possibilities have been overlooked, such as certain types of mesothelioma or metastatic tumors.

In order to further differentiate the diagnosis, the following additional examinations may be required:
\begin{itemize}
  \item \textbf{Immunohistochemistry}: such as TdT, p63, CD30, CD20, CK, Syn, and CgA, to distinguish thymoma, lymphoma, germ cell tumors, and neuroendocrine tumors.
  \item \textbf{Molecular testing}: such as KRAS mutation testing, NRAS mutation testing, and BRAF mutation testing, to further confirm germ cell tumors or other specific types of tumors.
\end{itemize}

At present, the differential diagnosis list includes thymic carcinoma / thymoma, lymphoma, germ cell tumor, and neuroendocrine tumor. These diagnoses are relatively likely, and the total number does not exceed ten.

The differential diagnoses and additional examination items are summarized as follows.

\textless /think\textgreater

\textless answer\textgreater

\textbf{Differential diagnosis:} 

\textbackslash DiffList\{\textcolor{green!60!black}{Thymic carcinoma} / thymoma, Lymphoma, Germ cell tumor, Neuroendocrine tumor\}

\vspace{0.5em}

\textbf{Further Examination Items:} 

\textbackslash ExamList\{Immunohistochemistry (TdT, p63, CD30, CD20, CK, Syn, CgA), Molecular testing (KRAS mutation testing, NRAS mutation testing, BRAF mutation testing)\}

\vspace{0.5em}

\textbf{Further Observation Tool Calls:} 

\textbackslash ToolCallList\{\}

\textless /answer\textgreater

\end{tcolorbox}

\begin{tcolorbox}[
    colback=gray!10,      
    colframe=black,       
    title=Additional Evidence (Round 1: only IHC),
    fonttitle=\bfseries\color{white},  
    coltitle=black,       
    boxrule=0.5pt,        
    arc=2mm,              
    top=2mm, bottom=2mm, left=2mm, right=2mm,
    breakable,
]

\textbf{Additional Test Results:} 

TdT: negative \qquad p63: positive \qquad CD30: negative \qquad CD20: negative \qquad CK: positive \qquad Syn: negative \qquad CgA: negative

\end{tcolorbox}

\begin{tcolorbox}[
    colback=gray!10,      
    colframe=black,       
    title=Diagnostic Reasoning (Second Round),
    fonttitle=\bfseries\color{white},  
    coltitle=black,       
    boxrule=0.5pt,        
    arc=2mm,              
    top=2mm, bottom=2mm, left=2mm, right=2mm,
    breakable,
]
\textless think\textgreater

First, the previously reviewed case information and the first round of diagnosis are revisited.

\textbf{Case information:}
\begin{itemize}
  \item The patient is a 47-year-old Asian female whose chief complaints are fatigue, weight loss, chest pain, and dyspnea.
  \item CT imaging shows an irregular, infiltrative mass in the anterior mediastinum, measuring approximately $8.3 \times 7.1 \times 8.2\,\mathrm{cm}$.
  \item The biopsy results show that the tissue is highly cellular, with tumor cells arranged in clusters and sheets. In some areas, glandular or papillary structures may be formed, but these are not obvious. The cells are relatively uniform in size and shape, with an increased nuclear-to-cytoplasmic ratio. The nuclear chromatin is finely distributed, and the nucleoli are small and inconspicuous. There is no obvious nuclear atypia or mitotic figures. The cells are arranged in solid sheets, and some areas show a loose or papillary growth pattern.
\end{itemize}

\textbf{First-round differential diagnosis:}
\begin{itemize}
  \item Thymic carcinoma / thymoma
  \item Lymphoma
  \item Germ cell tumor
  \item Neuroendocrine tumor
\end{itemize}

\textbf{Additional examination results:}
\begin{itemize}
  \item TdT: negative
  \item p63: positive
  \item CD30: negative
  \item CD20: negative
  \item CK: positive
  \item Syn: negative
  \item CgA: negative
\end{itemize}

According to the additional immunohistochemical results:
\begin{itemize}
  \item TdT negativity excludes the lymphocytic component of thymoma.
  \item p63 positivity suggests a possible association with thymic epithelial cells and supports thymic carcinoma / thymoma.
  \item CD30 negativity excludes embryonal carcinoma.
  \item CD20 negativity excludes B-cell derived lymphoma.
  \item CK positivity supports an epithelial origin and supports thymic carcinoma / thymoma.
  \item Syn and CgA negativity exclude neuroendocrine tumors.
\end{itemize}

Based on the above information, lymphoma, germ cell tumor, and neuroendocrine tumor can be preliminarily excluded, while thymic carcinoma / thymoma is more likely. Further differential diagnosis may require more detailed pathological features and histological subtype analysis.

\textbf{Final differential diagnosis:}
\begin{itemize}
  \item Thymic carcinoma / thymoma
\end{itemize}

\textbf{Additional examination items:}
\begin{itemize}
  \item In order to further distinguish thymic carcinoma from thymoma, additional immunohistochemical studies may be required, such as CD5, EMA, D2-40, and WT1.
\end{itemize}

The summary is as follows.

\textless /think\textgreater

\textless answer\textgreater

\textbf{Differential diagnosis:}  

\textbackslash DiffList\{\textcolor{green!60!black}{Thymic carcinoma} / thymoma\}

\vspace{0.5em}

\textbf{Further Examination Items:}  

\textbackslash ExamList\{Immunohistochemistry (CD5, EMA, D2-40, WT1)\}

\textless /answer\textgreater
\end{tcolorbox}

\begin{tcolorbox}[
    colback=gray!10,      
    colframe=black,       
    title=Additional Evidence (Round 2: extra IHC and molecular testing),
    fonttitle=\bfseries\color{white},  
    coltitle=black,       
    boxrule=0.5pt,        
    arc=2mm,              
    top=2mm, bottom=2mm, left=2mm, right=2mm,
    breakable,
]
\textbf{Additional examination results:}

CD5: positive \qquad 
EMA: positive \qquad 
D2-40: negative \qquad 
WT1: negative \qquad 
KRAS: no mutation detected \qquad 
NRAS: no mutation detected \qquad 
BRAF: no mutation detected \qquad 

\end{tcolorbox}

\begin{tcolorbox}[
    colback=gray!10,      
    colframe=black,       
    title=Diagnostic Reasoning (Final Decision),
    fonttitle=\bfseries\color{white},  
    coltitle=black,       
    boxrule=0.5pt,        
    arc=2mm,              
    top=2mm, bottom=2mm, left=2mm, right=2mm,
    breakable,
]
\textless think\textgreater

First, the previously reviewed case information and several rounds of diagnosis are revisited.

\textbf{Case information:}
\begin{itemize}
  \item The patient is a 47-year-old Asian female whose chief complaints are fatigue, weight loss, chest pain, and dyspnea.
  \item CT imaging shows an irregular, infiltrative mass in the anterior mediastinum, measuring approximately $8.3 \times 7.1 \times 8.2\,\mathrm{cm}$.
  \item The biopsy results show that the tissue is highly cellular, with tumor cells arranged in clusters and sheets. In some areas, glandular or papillary structures may be formed, but these are not obvious. The cells are relatively uniform in size and shape, with an increased nuclear-to-cytoplasmic ratio. The nuclear chromatin is finely distributed, and the nucleoli are small and inconspicuous. There is no obvious nuclear atypia or mitotic figures. The cells are arranged in solid sheets, and some areas show a loose or papillary growth pattern.
\end{itemize}

\textbf{First-round differential diagnosis:}
\begin{itemize}
  \item Thymic carcinoma / thymoma
  \item Lymphoma
  \item Germ cell tumor
  \item Neuroendocrine tumor
\end{itemize}

\textbf{Additional examination results (immunohistochemistry):}
\begin{itemize}
  \item TdT: negative
  \item p63: positive
  \item CD30: negative
  \item CD20: negative
  \item CK: positive
  \item Syn: negative
  \item CgA: negative
\end{itemize}

\textbf{Additional examination results (immunohistochemistry and molecular testing):}
\begin{itemize}
  \item CD5: positive
  \item EMA: positive
  \item D2-40: negative
  \item WT1: negative
  \item KRAS: no mutation
  \item NRAS: no mutation
  \item BRAF: no mutation
\end{itemize}

According to the additional immunohistochemical and molecular testing results:
\begin{itemize}
  \item TdT negativity excludes the lymphocytic component in thymoma.
  \item p63 positivity suggests a possible association with thymic epithelial cells and supports thymic carcinoma / thymoma.
  \item CD30 negativity excludes embryonal carcinoma.
  \item CD20 negativity excludes B-cell derived lymphoma.
  \item CK positivity supports an epithelial origin and supports thymic carcinoma / thymoma.
  \item Syn and CgA negativity exclude neuroendocrine tumors.
  \item CD5 positivity supports thymic carcinoma.
  \item EMA positivity further supports a thymic epithelial cell origin.
  \item D2-40 negativity excludes lesions of pleural origin.
  \item WT1 negativity excludes certain types of thymoma.
  \item No mutations in KRAS, NRAS, and BRAF exclude certain types of germ cell tumors.
\end{itemize}

Based on the above information, the diagnosis of thymic carcinoma / thymoma is further supported, with thymic carcinoma being more likely.

\textbf{Final diagnosis:}
\begin{itemize}
  \item Thymic carcinoma
\end{itemize}

The summary is as follows.

\textless /think\textgreater

\textless answer\textgreater

\textbackslash boxed\{\textcolor{green!60!black}{Thymic carcinoma}\}

\textless /answer\textgreater

\end{tcolorbox}

This case involves the evaluation of a biopsy specimen. For pathologists, establishing a diagnosis based on such a small biopsy specimen is a significant challenge. The incorporation of models can relieve the burden by drafting initial microscopic findings, suggesting possible diagnostic directions, and proposing further examinations. However, it remains a considerable space for automatic models. For example, molecular testing is redundant in this case, and the conclusion can be more precise if it reads ``A tumor of thymic epithelial origin, favoring thymic carcinoma,'' since it is only a biopsy. Although PathFound needs improvement to become an experienced expert in pathology, it deserves to be a decent agentic copilot for pathologists.

\section{Declaration of generative AI and AI-assisted technologies in the manuscript preparation process}
During the preparation of this work the authors used GPT-5 in order to improve the readability of the manuscript. After using this tool/service, the authors reviewed and edited the content as needed and take full responsibility for the content of the published article.

%%       \citet{<label>} ==> Lamport (1994)
%%      ~\citep{<label>} ==> (Lamport, 1994)
%%
% Example citation, See \citet{lamport94}.

\bibliographystyle{elsarticle-harv} 
\bibliography{reference}




% %% For authoryear reference style

% \bibitem[Lamport(1994)]{lamport94}
%   \textit{\LaTeX: a document preparation system},
%   Addison Wesley, Massachusetts,

% \end{thebibliography}
\end{document}